\documentclass[11pt,a4paper]{article}

\usepackage{polyu-vclab}

\usepackage{lipsum}                
\usepackage{amsmath}
\usepackage{booktabs}
\usepackage{multirow}
\usepackage[numbers,sort&compress]{natbib}
\usepackage{graphicx}
\usepackage{cleveref}
\usepackage{fvextra}
\usepackage{adjustbox}
\usepackage{enumitem}

\setEyebrow{PolyU VCLab\,\textbullet\,Preprint 2026}

\setReportTag{Visual Computing Lab\,\textperiodcentered\,The Hong Kong Polytechnic University}

\papertitle{Avatar-Forever: Decoupled Parallel Training for High-Quality Real-Time Infinite Avatars}

\paperauthors{%
  Ruibin Li\affilmark{1}\equalmark\quad
  Zhiyuan Ma\affilmark{1}\quad
  Fangzhou Ai\affilmark{3}\quad
  Lei Zhang\correspondmark\affilmark{1}%
}

\paperaffil{%
  \affilmark{1}\,The Hong Kong Polytechnic University\quad
  \affilmark{2}\,ByteDance\quad
  \affilmark{3}\,AMD\quad
}

\papernotes{%
  \equalmark\,Work done during internship in ByteDance.\quad
  \correspondmark\,Corresponding author (\href{mailto:cslzhang@comp.polyu.edu.hk}{cslzhang@comp.polyu.edu.hk}).%
}

\paperbadges{%
  \vclabbadgesolid{Project Page}{https://leeruibin.github.io/avatarforever-project-page/}\;%
  \vclabbadgesolid{Code}{https://github.com/leeruibin/avatarforever/tree/main}\;%
  \vclabbadgesolid{Demo}{https://leeruibin.github.io/avatarforever-project-page/}\;%
  \vclabbadgesolid{BibTeX}{https://polyu-vclab.github.io/cite}%
}

\begin{document}

\maketitleVCLab


\begin{vclabAbstract}
\noindent\textbf{Abstract.}\;
Existing streaming video systems often rely on sequential, distillation-centered training pipelines to enable few-step long-video generation. 
However, this paradigm suffers from two limitations. 
First, failures or distribution shifts introduced in earlier stages affect later optimization, complicating the training process to converge. 
Second, the distillation-centric objective favours short-term generation but is prone to quality degradation when autoregressive errors accumulate over long rollouts. 
We propose \emph{Avatar-Forever}, a decoupled parallel training framework for high-quality real-time infinite interactive avatars. Instead of coupling generation efficiency and long-horizon robustness under a sequential distillation pipeline, we treat them as two independent capabilities that can be trained in parallel. One branch performs full-parameter distillation to train an efficient generator with high visual quality, while another trains a lightweight long-horizon adapter via \emph{Recovery-oriented Rollout Training} (RRT), which improves generation robustness under long-horizon inference conditions. Our decoupled parallel training design simplifies the overall training process and avoids unnecessary objective conflicts between few-step generation and long-horizon adaptation. 
We further introduce \emph{ForeverCache}, a chunk-wise feature caching mechanism to substantially reduce redundant history computation during streaming inference. 
Built upon a 22B video foundation model, Avatar-Forever supports unbounded audio-driven avatar generation while maintaining identity consistency, motion coherence, and visual fidelity, enabling an end-to-end throughput of high-resolution $768\times512$ videos at 27.2 FPS on a single H100 GPU and providing a practical path toward stable digital humans.

\end{vclabAbstract}


\section{Introduction}

\begin{figure}[t!]
\centering
\includegraphics[width=\textwidth]{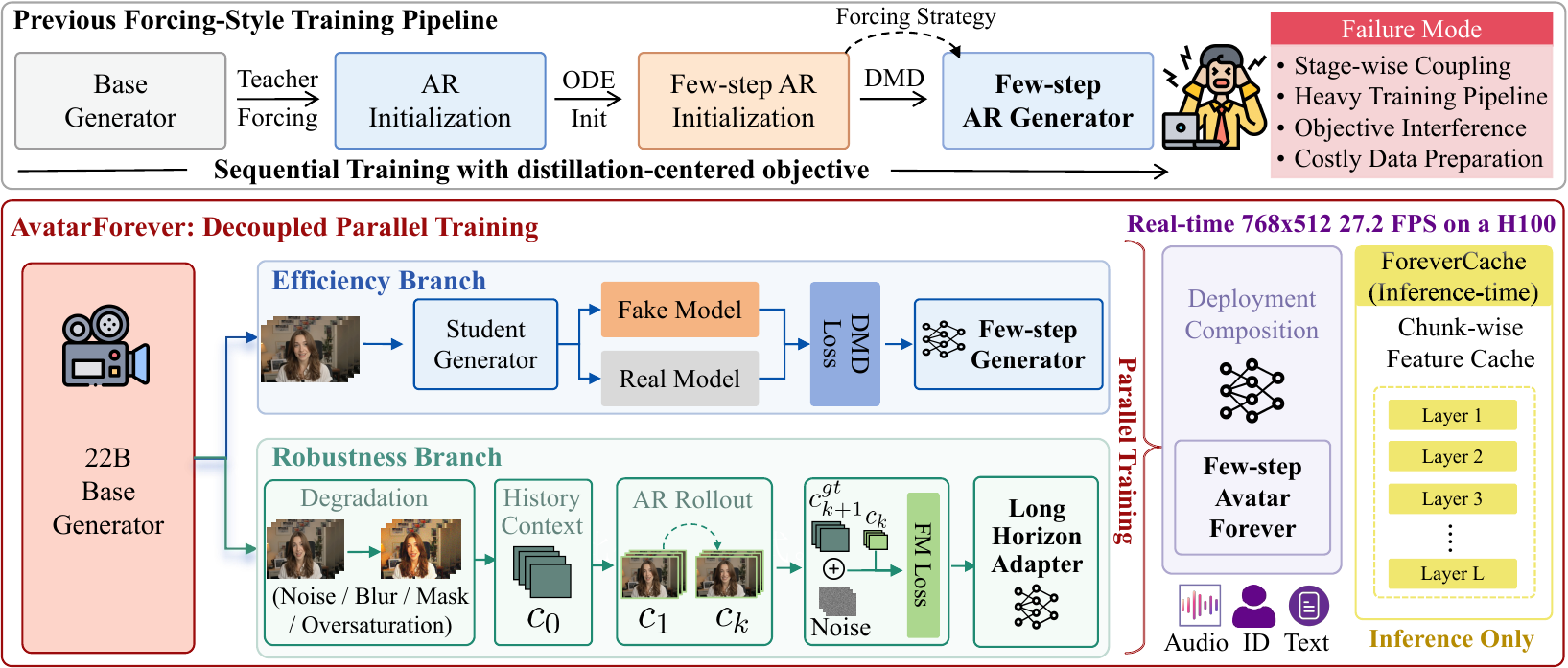}
\caption{\textbf{Overview of Avatar-Forever.} \textit{Top:} Existing streaming avatar systems employ sequential distillation pipelines, leading to stage-wise dependence and objective interference. \textit{Bottom:} Avatar-Forever learns few-step efficiency and long-horizon robustness in parallel. A lightweight adapter is trained under accumulated rollout errors, and it is merged with the distilled generator in inference. In addition, ForeverCache is proposed to reuse historical features for efficient streaming inference.}
\centering
\label{fig:overview}
\end{figure}

Generating realistic and expressive digital humans is highly demanded in applications such as virtual assistants, content creation, education, and entertainment~\cite{zhang2023sadtalker,li2024latentsync,tian2024emo}. 
A practical avatar system should continuously synthesize coherent video over extended durations while maintaining identity consistency, audio-driven motion, visual fidelity, and efficient inference.
Recent diffusion-based video foundation models~\cite{seedance2026seedance2,wan2025wan,hacohen2026ltx,li2025many,davinci-magihuman-2026} have significantly advanced audio-driven avatar generation, producing photorealistic talking avatars with improved visual fidelity, motion realism, and lip synchronization~\cite{guo2025arig,li2025ditto,low2025talkingmachines,cui2026avatarforcing}. However, most models are trained to generate a short clip within a fixed temporal window, and extending these advances from short clips to real-time long-duration avatar generation remains challenging~\cite{yin2025slow,huang2026self,sun2026streamavatar}. 
To generate longer outputs, streaming systems typically reuse previously generated chunks as conditions for future chunks~\cite{yang2025longlive,lu2026reward,li2025stable,zhu2026causalforcing,yesiltepe2026infinity}, which introduces distribution drift and error accumulation: early artifacts, identity drift, or lip-sync errors become part of the historical context and gradually degrade subsequent generation~\cite{yuan2026helios,chen2026longlive}.

Recent works \cite{zeng2026lpm,shen2026soulxflashtalk,sun2026streamavatar} usually address this challenge with distillation-centered sequential training pipelines, as illustrated in the top of Figure~\ref{fig:overview}. Starting from a multi-step bidirectional video model, these methods combine teacher forcing, ODE initialization, DMD distillation, and additional forcing strategies~\cite{yin2025slow,huang2026self,zhu2026causalforcing,cui2025selfforcingpp,yang2025infinitetalk,huang2025liveavatar} with supervision imposed in fixed denoising steps to obtain a few-step model. This sequential design entangles two distinct objectives in the optimization pipeline: efficient few-step generation, which preserves short-term visual quality under reduced denoising steps, and long-horizon robustness, which stabilizes generation against accumulated autoregressive errors. Such an entanglement makes the training fragile and difficult to scale: distribution shifts introduced at earlier stages affect later optimization, making it hard to diagnose and obscuring the contribution of each stage.

We argue that the two objectives of efficient few-step generation and long-horizon robustness should be decoupled and learned in parallel, and propose \emph{Avatar-Forever}, a decoupled parallel training framework for high-quality real-time long-duration audio-driven avatars, as illustrated in the bottom of Figure~\ref{fig:overview}. Instead of intertwining distillation, forcing schedules and long-context adaptation in a sequential pipeline, we train for efficiency and robustness in parallel. One branch performs full-parameter distillation to obtain a high-quality few-step generator, while the other trains a lightweight long-horizon adapter for stability under extended rollout conditions. This design simplifies the training procedure, making each objective easier to optimize and diagnosis. For the long-horizon adaptation branch, we introduce \emph{Recovery-oriented Rollout Training} (RRT), which targets the main failure mode of autoregressive generation, \emph{i.e.}, errors accumulate after recursive rollout. Starting from a degraded historical context that simulates generation artifacts, the model rolls out multiple future chunks using its own predictions as conditions. Supervision is applied when the degradation has propagated through the rollout, rather than at every local reconstruction step. 
Importantly, RRT uses a standard flow matching objective and does not require specialized long-video distillation losses or complicated regularization schemes.

During streaming inference, each new chunk needs to attend to historical chunks. One simple way is to recompute the entire visible history window at every denoising step, although only the current chunk is updated. We introduce \emph{ForeverCache}, a chunk-wise autoregressive history feature caching mechanism to remove redundant computation. For each generated chunk, we perform a full forward pass at the first denoising step to populate per-block historical context features, and then reuse these stable features in subsequent denoising steps while forwarding only the current chunk tokens. This inference-time mechanism preserves the historical context needed for identity consistency, motion continuity, and audio-driven synchronization, while improving throughput by \textbf{23\%} during long-horizon generation.

To support scalable adaptation, we construct a fully synthetic data pipeline by using video foundation models to synthesize training videos from high-quality conversational prompts, followed by automatic filtering to remove low-quality, static, or semantically inconsistent samples. Built upon a 22B video foundation model \cite{hacohen2026ltx} and our synthetic dataset, Avatar-Forever scales audio-driven avatar generation to extended durations while maintaining visual fidelity, identity consistency, and motion coherence. 
Specifically, Avatar-Forever achieves an \textbf{end-to-end} (including DiT inference and VAE decoding) generation throughput of \textbf{27.2 FPS} for \textbf{high-quality $\mathbf{768\times512}$} avatar videos on a single H100 GPU, achieving state-of-the-art performance on long-horizon audio-driven avatar generation.
Our main contributions are summarized as follows:

\begin{itemize}

    \item We propose \textbf{Avatar-Forever}, a decoupled parallel training paradigm for real-time long-duration avatar generation,
     showing that few-step efficiency and autoregressive robustness can be effectively learned in parallel, rather than in a complicated sequential manner. 

    \item We introduce \textbf{Recovery-oriented Rollout Training} (RRT) in order to recover from errors after propagating through multi-chunk autoregressive rollout. Unlike local corrupted-context reconstruction or long-video distillation objectives, RRT uses standard flow matching supervision under the rollout distribution encountered at inference time.

    \item We introduce \textbf{ForeverCache}, a chunk-wise history feature cache that reuses stable context features across denoising steps, removing redundant recomputation of  historical chunks during streaming inference and improving long-horizon generation throughput by \textbf{23\%}.

    \item We implement the framework with a fully synthetic data pipeline and a 22B video foundation model, achieving high-quality long-duration audio-driven avatar generation with persistent identity, coherent motion, accurate lip synchronization, and real-time $\mathbf{768\times512}$ resolution inference at \textbf{27.2 FPS} on a single H100 GPU.

\end{itemize}

\begin{vclabNote}
\textbf{Remarks.}\; Few-step efficiency and long-horizon stability operate on different temporal scales. Coupling them as one distillation objective entangles speed with robustness, complicating the training pipeline. Avatar-Forever learns to achieve them in a parallel manner, significantly improving the training process.
\end{vclabNote}

\section{Related Work}
\label{sec:related_work}

\subsection{Audio-Driven Avatar Video Generation}

\paragraph{Traditional and 3D-based Methods.}
Audio-driven avatar generation has been widely studied through 2D talking-head synthesis \cite{pham2024style}, explicit facial representations \cite{deng2020disentangled}, and neural rendering \cite{cudeiro2019capture}. Early systems primarily focus on accurate lip synchronization and identity preservation from limited visual context~\citep{prajwal2020wav2lip}. To improve structural stability, later methods introduce 3D priors \cite{cao2024dreamavatar}, facial motion coefficients \cite{zhang2023sadtalker}, or neural radiance fields \cite{xu2023latentavatar}, mapping speech to geometrically meaningful facial dynamics before rendering. These methods provide strong priors for lip motion, head pose, and identity consistency, but they are basically task-specific generators or subject-specific renders, offering limited flexibility to diverse identities, expressive body motion, complex scenes, and open-ended behaviors \cite{huang2025liveavatar}. Their reliance on predefined structures or identity-specific rendering also limits the generalization to unconstrained avatar scenarios, especially for long-duration generation.

\noindent\textbf{Diffusion-based Avatars.}
Diffusion models \cite{ho2020denoising,lipman2022flow,rombach2022high} have recently become increasingly important backbones for avatar synthesis, as they can capture subtle facial dynamics, photorealistic appearance, and temporal variations. Task-specific avatar systems such as DiffTalk \cite{shen2023difftalk}, EMO \cite{tian2024emo}, Hallo \cite{xu2024hallo}, AniPortrait \cite{wei2024aniportrait} and EchoMimic \cite{chen2024echomimic} adapt diffusion models to audio-driven portrait animation through audio-conditioned denoising, hierarchical control, or editable motion guidance. Meanwhile, large video foundation models, exemplified by Wan \cite{wan2025wan}, Seedance 2.0 \cite{seedance2026seedance2}, Kling-Omni \cite{kling2025omni}, and Sora \cite{openai2024sora}, have demonstrated that scaled diffusion-transformer video generators can serve as strong spatiotemporal priors, although they are not specifically designed for audio-driven avatar generation. However, these models introduce substantially high inference cost, making fast sampling a critical bottleneck. While diffusion acceleration methods have been developed to reduce the denoising steps~\citep{yin2024dmd,yin2024dmd2,sauer2024add,luo2023lcm,wang2024pcm,wu2026diversity}, they mainly address the problem of efficient short-horizon generation, whereas long-duration avatar synthesis also requires robustness under recursive autoregressive rollout.

\subsection{Long-Horizon and Streaming Video Diffusion Models}

Streaming video generation~\cite{huang2025selfforcing,zhu2026causalforcing,huang2025liveavatar,li2026hybridforcing} reuse previously generated frames as conditions for future predictions to conduct long-horizon video generation. However, this creates a train--test mismatch: models are trained on clean ground-truth contexts but must condition on their own imperfect outputs at inference, causing artifacts, motion drift, and temporal inconsistency to accumulate over time~\cite{shen2026soulxflashtalk,yuan2026helios,zeng2026lpm}. To reduce the train--test mismatch in autoregressive long-video generation, recent methods expose models to self-generated contexts during training or distillation. Self-Forcing++~\cite{cui2025selfforcingpp}, Causal Forcing~\cite{zhu2026causalforcing}, Rolling Forcing~\cite{liu2025rollingforcing}, and Hybrid Forcing~\cite{li2026hybridforcing} introduce mechanisms such as teacher-guided rollout, auto-regressive teacher initialization, progressive denoising, and hybrid attention design to improve long-horizon stability. These forcing strategies substantially improve the stability and quality of streaming generation. However, they often rely on tightly coupled multi-stage pipelines, making component-wise diagnosis difficult and scaling to large video foundation models cumbersome.

Long-duration avatar generation faces the same autoregressive streaming challenge, but with stricter requirements on identity preservation, appearance stability, lip synchronization, facial dynamics, and human motion coherence~\cite{zeng2026lpm,gan2025omniavatar,chen2025hunyuanvideo,davinci-magihuman-2026}. Recent systems such as LiveAvatar~\cite{huang2025liveavatar}, StreamAvatar~\cite{sun2026streamavatar}, SoulX-FlashTalk~\cite{shen2026soulxflashtalk}, and LPM~1.0~\cite{zeng2026lpm} adapt large video priors to streaming or long-horizon human synthesis, showing the potential of video foundation models for interactive digital humans. However, these methods largely follow the sequential training paradigm of generic long-video generation, where multiple stages are coupled through autoregressive rollout, distillation, and complicated foring strategies. As a result, errors and distribution shifts~\cite{cui2025selfforcingpp,yuan2026helios} introduced in early stages can propagate through the pipeline, making the final behavior hard to diagnose and costly to scale to large video foundation models.

\section{Method}

As illustrated in the bottom of Figure~\ref{fig:overview}, Avatar-Forever consists of three core designs. An \textit{efficiency branch} and a \textit{robustness branch} are trained separately via decoupled adaptation. The full-parameter efficiency branch is distilled to preserve high-quality few-step generation, while the robust branch is adapted to improve long-horizon stability. The resulting long-horizon LoRA adapter is merged into the distilled generator at inference. Within the robustness branch, we propose \emph{Recovery-oriented Rollout Training} (RRT), which perturbs early historical context, propagates the degradation through multi-step autoregressive rollout, and applies standard flow-matching supervision only after errors have accumulated. 
In addition, a \emph{ForeverCache} strategy is proposed to accelerate streaming inference by caching stable historical context features across denoising steps, avoiding redundant recomputation of clean history chunks while preserving long-range conditioning.



\subsection{Efficiency Branch: Few-Step Distillation}
\label{sec:distillation_branch}

The efficiency branch aims to obtain a high-quality few-step generator from the pretrained video foundation model. We adopt Distribution Matching Distillation (DMD)~\cite{yin2024dmd2} to compress the original multi-step LTX base model~\cite{hacohen2026ltx} into a four-step generator while preserving conditional generation quality. Let $\theta_0$ denote the pretrained model parameters, $G_\theta$ denote the student generator, and $\mathbf{c}$ denote the conditioning inputs, including text and optional visual conditions. DMD optimizes the student by matching the generated distribution to the teacher distribution through a reverse-KL objective, formulated as follows:
\begin{equation}
\label{eq:dmd-gradient}
\nabla_{\theta}\mathcal{L}_{\mathrm{DMD}}
=
-
\mathbb{E}_{t,\boldsymbol{\epsilon},\mathbf{c}}
\left[
\left(
s_{\mathrm{real}}(\tilde{\mathbf{x}}_t)
-
s_{\mathrm{fake}}(\tilde{\mathbf{x}}_t)
\right)
\frac{\partial G_\theta(\boldsymbol{\epsilon},\mathbf{c})}{\partial \theta}
\right],
\end{equation}
where $\tilde{\mathbf{x}}_t$ is obtained by applying the forward diffusion process to the student sample $G_\theta(\boldsymbol{\epsilon},\mathbf{c})$, and $s_{\mathrm{real}}$ and $s_{\mathrm{fake}}$ are score functions. This branch is responsible only for few-step efficiency. We do not introduce autoregressive rollout, corrupted history, or long-horizon objectives during distillation. To preserve compatibility with downstream avatar generation, we train the distilled model under mixed conditioning: each sample is randomly used either as a text-to-video instance or as a first-frame-conditioned instance, where the first frame serves as the visual condition. This preserves both T2V and I2V capabilities of the base model while keeping the distillation objective focused on short-horizon quality, motion realism, and efficient sampling.

\subsection{Robustness Branch: Recovery-oriented Rollout Training}
\label{sec:rrt}

\begin{figure}[t!]
\centering
\includegraphics[width=\textwidth]{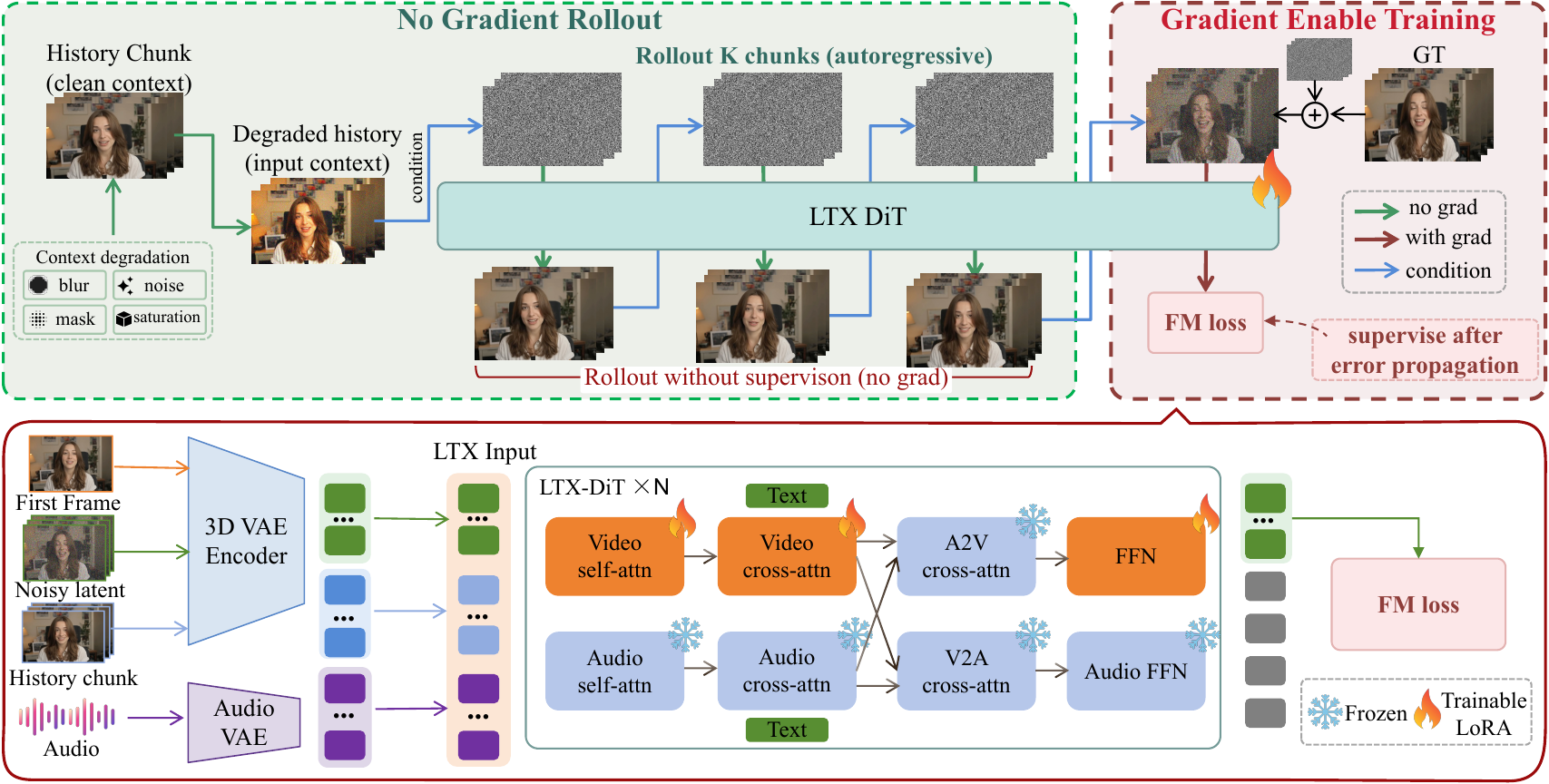}
\caption{\textbf{Recovery-oriented Rollout Training (RRT).} \textit{Top:} RRT degrades an early history chunk, propagates the resulting error through $K$ autoregressive rollout steps without gradients, and applies standard FM supervision only to the subsequent recovery step. \textit{Bottom:} In LTX DiT, video and audio conditions are encoded into multimodal tokens; the gated first-frame pathway and video-side LoRA modules are trained, while the remaining components are frozen.}
\centering
\label{fig:detail_rrt}
\end{figure}

The robustness branch addresses the train--test mismatch introduced by autoregressive self-conditioning. At inference time, generated chunks become historical context for future chunks, so early errors can be recursively reused and amplified, causing identity drift, motion incoherence, and appearance artifacts~\cite{yuan2026helios,shen2026soulxflashtalk}. Training only on clean histories may lead to failures when exposed to deployment-time error distribution. A natural remedy is corrupted-history training, as explored in Helios~\cite{yuan2026helios}. However, local corrupted-context reconstruction only approximates an isolated perturbation, while long-horizon failure arises from accumulated self-generated drift. We therefore introduce \emph{Recovery-oriented Rollout Training} (RRT): we perturb an early history chunk, let the model propagate this perturbation through multi-step autoregressive rollout, and apply supervision after errors have accumulated. RRT thus trains recovery under model-induced context drift rather than one-step synthetic corruption. The details of RRT are illustrated in Figure~\ref{fig:detail_rrt}.

\noindent\textbf{Early-history Perturbation.}
We partition each training video latent $\mathbf{x}_0$ into chunks $\{\mathbf{c}_k\}_{k=0}^{K+1}$, where $\mathbf{c}_0$ denotes the initial context, $\mathbf{c}_{1:K}$ are intermediate rollout chunks, and $\mathbf{c}_{K+1}$ is the final supervised target chunk. We construct the initial perturbed history as:
\begin{equation}
    \hat{\mathbf{c}}_0 = \mathcal{D}(\mathbf{c}_0),
\end{equation}
where $\mathcal{D}(\cdot)$ is a stochastic degradation operator. Inspired by Helios~\cite{yuan2026helios}, $\mathcal{D}$ randomly samples from a lightweight family of perturbations, including photometric distortion, additive noise, resolution degradation, partial latent masking, and identity mapping. Importantly, the perturbation is applied only to the earliest conditioning history. All subsequent histories are generated by the model itself and recursively reused as conditions. The initial perturbation therefore serves as a trigger for distribution shift, while later errors emerge from the model's own rollout dynamics.

\noindent\textbf{Autoregressive Rollout.}
Let $\mathbf{a}_k$ denote the audio condition aligned with chunk $\mathbf{c}_k$, and let $y$ denote the text condition. For each intermediate chunk $k=1,\ldots,K$, we initialize from Gaussian noise $\hat{\mathbf{c}}_{k,T}\sim\mathcal{N}(\mathbf{0},\mathbf{I})$ and perform $T$ denoising steps conditioned on the previous generated chunk:
\begin{equation}
\label{eq:rrt-rollout}
\hat{\mathbf{c}}_{k,t}
=
\operatorname{sg}
\left(
G_{\theta}
\left(
\hat{\mathbf{c}}_{k,t+1};
\hat{\mathbf{c}}_{k-1,0}, \mathbf{a}_k, y
\right)
\right),
\quad
t=T-1,\ldots,0,
\end{equation}
where $\hat{\mathbf{c}}_{k-1,0}$ is the generated result of the previous chunk, $G_{\theta}$ denotes one denoising step of the sampler, and $\operatorname{sg}(\cdot)$ stops gradients through the rollout trajectory. As illustrated in the top left of Figure~\ref{fig:detail_rrt}, rollout is executed without gradient tracking. The intermediate chunks $\hat{\mathbf{c}}_1,\ldots,\hat{\mathbf{c}}_K$ are not directly supervised; their role is to expose the final prediction to model-induced context drift, so that optimization targets recovery under the same autoregressive error-propagation pattern encountered at inference time.

\noindent\textbf{Masked Flow Matching Objective.}
After $K$ rollout steps, we supervise only the next ground-truth chunk $\mathbf{c}_{K+1}$, conditioned on the generated history $\hat{\mathbf{c}}_{K,0}$. Given noise $\boldsymbol{\epsilon}\sim\mathcal{N}(\mathbf{0},\mathbf{I})$ and a flow-matching noise level $\sigma\sim p(\sigma)$, we corrupt only the target tokens:
\begin{equation}
\label{eq:rrt-forward}
    \hat{\mathbf{c}}_{K+1,\sigma}
    = (1-\sigma)\mathbf{c}_{K+1} + \sigma\boldsymbol{\epsilon},
\end{equation}
while keeping the rolled-out historical context $\hat{\mathbf{c}}_{K,0}$ clean. We then optimize a simple flow-matching~\cite{lipman2022flow} objective on the final target chunk:
\begin{equation}
\label{eq:rrt-loss}
    \mathcal{L}_{\mathrm{RRT}}
    =
    \mathbb{E}_{\mathbf{c},\mathbf{a},y,\sigma,\boldsymbol{\epsilon}}
    \left[
        \left\|
            v_{\theta}
            \left(
                \mathbf{c}_{K+1,\sigma}, \sigma;
                \hat{\mathbf{c}}_{K,0}, \mathbf{a}_{K:K+1}, y
            \right)_i
            -
            \left(
                \boldsymbol{\epsilon} - \mathbf{c}_{K+1}
            \right)_i
        \right\|_2^2
    \right],
\end{equation}
where $v_\theta$ is the velocity network, $\mathbf{a}_{K:K+1}$ denotes the audio condition aligned with the visible context-target window. We do not impose auxiliary losses on the intermediate rollout chunks. Once realistic autoregressive degradation is induced through model-in-the-loop rollout, a standard flow-matching objective on the final target chunk is sufficient to train recovery from accumulated long-horizon errors.

\begin{vclabNote}
\textbf{Insight.}\; We do not supervise immediate local reconstruction from a degraded context; we supervise recovery after degradation has propagated through autoregressive rollout.
\end{vclabNote}

\noindent\textbf{Deployment}.
Both adaptation branches are initialized from the same pretrained base model $\theta_0$ so that their updates can be merged directly at inference. Let $\Delta\theta_{\mathrm{DMD}}$ denote the dense update learned by the efficiency branch, and let $\Delta\theta_{\mathrm{RRT}}$ denote the LoRA update learned by the robustness branch. The final Avatar-Forever generator is obtained as
\begin{equation}
    \theta^{\star}
    =
    \theta_0
    +
    \Delta\theta_{\mathrm{DMD}}
    +
    \Delta\theta_{\mathrm{RRT}}.
\end{equation}
This composition keeps the two objectives decoupled during training: full-parameter distillation provides a strong few-step generator, while the LoRA-based RRT branch adds long-horizon robustness with minimal additional training and deployment cost.

\subsection{ForeverCache: Chunk-wise Autoregressive History Feature Caching}
\label{sec:forevercache}

\begin{figure}[t!]
\centering
\includegraphics[width=\textwidth]{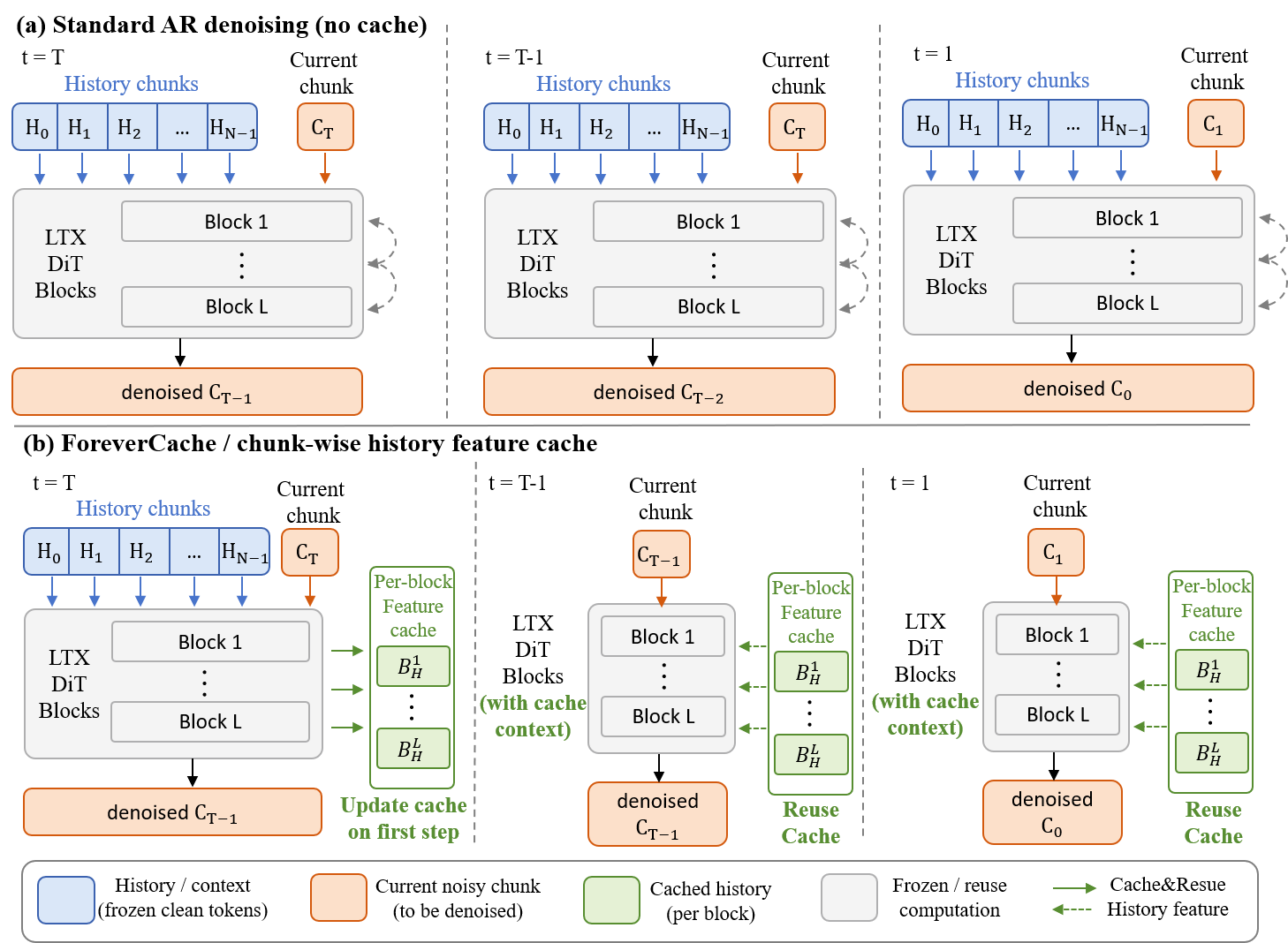}
\caption{\textbf{ForeverCache for autoregressive streaming inference.} \textit{Top:} Standard autoregressive denoising recomputes fixed history features at every step. \textit{Bottom:} ForeverCache populates per-block history features once at $t=T$ and reuses them while forwarding only the evolving current chunk thereafter. The cache is reset for each new autoregressive chunk, providing bounded-memory streaming inference without redundant history computation.}
\centering
\label{fig:forevercache}
\end{figure}

When producing chunk $k$, the autoregressive model denoises the current latent $\mathbf{c}_{k,t}$ while attending to a compact history window
$\mathcal{H}_k=\{\mathbf{c}_0,\mathbf{c}_{k-1}\}$,
consisting of the first and most recent generated chunks. A naive implementation forwards the entire visible window $\{\mathcal{H}_k,\mathbf{c}_{k,t}\}$ through every transformer block at each denoising step, as shown in Figure~\ref{fig:forevercache}(a). However, this computation is structurally redundant: $\mathbf{c}_{k,t}$ changes throughout denoising, whereas the historical chunks remain fixed clean context.

We introduce \emph{ForeverCache}, an inference-time feature reuse mechanism that eliminates repeated computation on fixed historical context. For each autoregressive chunk, ForeverCache executes a full-window forward pass only at the initial denoising step, then reuses the resulting historical features throughout the remaining denoising trajectory. This design retains access to long-range context while restricting repeated computation to the evolving current chunk, as illustrated in Figure~\ref{fig:forevercache}(b).

Formally, let $\sigma_t$ denote the noise level at denoising step $t$. A standard autoregressive implementation evaluates the velocity network on the complete visible window at every step:
\begin{equation}
    \mathbf{v}_{k,t}
    =
    v_\theta
    \left(
        [\mathcal{H}_k,\mathbf{c}_{k,t}],
        \sigma_t;
        \mathbf{a}_{\mathcal{H}_k:k}, y
    \right),
\end{equation}
where $[\mathcal{H}_k,\mathbf{c}_{k,t}]$ is the history-current sequence input to LTX~\cite{hacohen2026ltx}. Only the velocity prediction for the current chunk is retained for updating $\mathbf{c}_{k,t}$; the historical chunks act solely as conditioning context.

ForeverCache performs full computation only once at $t=T$, while collecting historical features from all the $L$ transformer blocks:
\begin{equation}
    \mathbf{v}_{k,T}, \mathcal{C}_k
    =
    v_\theta^{\mathrm{populate}}
    \left(
        [\mathcal{H}_k,\mathbf{c}_{k,T}],
        \sigma_T;
        \mathbf{a}_{\mathcal{H}_k:k}, y
    \right),
    \qquad
    \mathcal{C}_k=\{\mathcal{C}_k^\ell\}_{\ell=1}^{L}.
\end{equation}
For all subsequent steps, the model forwards only the current-chunk tokens and retrieves $\mathcal{C}_k$ as historical conditioning memory:
\begin{equation}
    \mathbf{v}_{k,t}
    =
    v_\theta^{\mathrm{reuse}}
    \left(
        \mathbf{c}_{k,t},
        \sigma_t;
        \mathcal{C}_k,
        \mathbf{a}_{k}, y
    \right),
    \qquad t=T-1,\ldots,0.
\end{equation}
Here, $v_\theta^{\mathrm{populate}}$ and $v_\theta^{\mathrm{reuse}}$ share the same model parameters but differ in their execution mode. At each transformer block, only current tokens are processed as active tokens, while cached features provide historical context for video self-attention, audio self-attention, and cross-modal audio-video attention. The resulting prediction is scattered back to the original autoregressive layout, preserving the external denoising interface of the non-cached implementation. The cache is reset for each autoregressive chunk, ensuring bounded memory and preventing stale reuse as the history window changes. After the initial cache-population step, ForeverCache reduces the dominant transformer cost by forwarding only the evolving current-chunk tokens, while reusing cached video and audio history features to preserve multimodal conditioning for identity consistency, motion continuity, and lip synchronization. Note that ForeverCache is used only at inference time and applies directly to the distilled Avatar-Forever generator without modifying any learned weights.

\section{Synthetic Data Construction Pipeline}
\label{sec:data_pipeline}

\begin{figure}[t!]
\centering
\includegraphics[width=\textwidth]{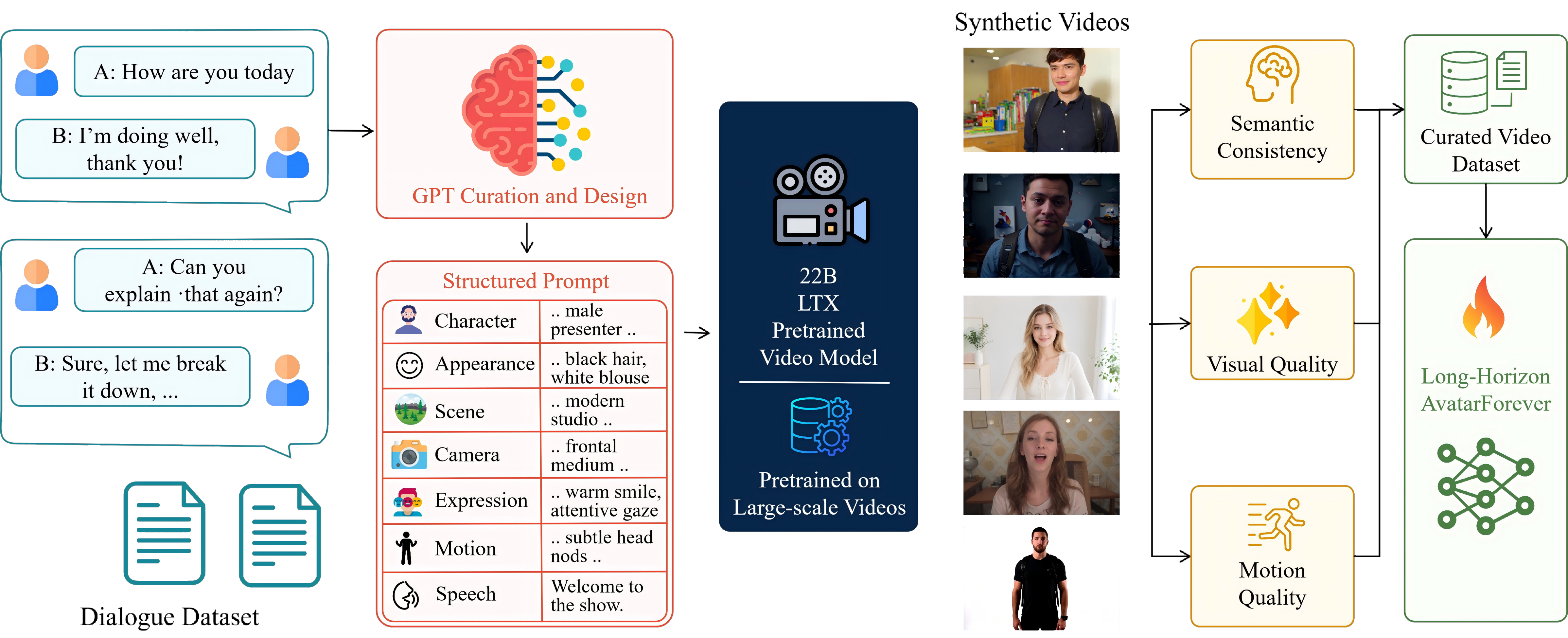}
\caption{\textbf{Synthetic data construction pipeline for long-horizon avatar adaptation.} Public dialogues are filtered and converted into structured video prompts, from which the pretrained LTX model synthesizes controllable avatar videos. Quality-aware filtering removes semantically mismatched, visually degraded, static, and camera-dominated samples, yielding a curated synthetic corpus for long-horizon avatar adaptation.}
\centering
\label{fig:foreverdata}
\end{figure}

Long-horizon avatar adaptation requires videos with persistent identity, meaningful human motion, and reliable semantic alignment over extended durations. Such data are difficult to collect from the web: real videos often exhibit identity changes, weak text--video correspondence, limited facial or body motion, and inconsistent visual quality. We address this limitation with a fully synthetic pipeline that generates and filters avatar videos using the same video foundation model adapted by our framework. This construction pipeline provides controllable training data that are naturally aligned with the target generation distribution, without large-scale manual curation of long-form real videos.

\noindent\textbf{Dialogue and Prompt Synthesis.}
We collect conversational text from the publicly available MDD corpus~\cite{dodge2016evaluating} and use GPT~\cite{openai2026chatgpt} to filter and refine samples that are incoherent, low-quality, or visually uninformative. The retained dialogues are converted into structured video prompts following the LTX prompting format. Each prompt specifies the character, appearance, scene, camera view, facial expression, body motion, and speech content, encouraging visible articulation and interaction-relevant motion in the resulting videos, as illustrated in the red part of Figure~\ref{fig:foreverdata}.

\noindent\textbf{LTX Video Synthesis.}
We feed the synthesized prompts to the pretrained LTX base model and generate avatar videos with standard multi-step sampling. Unlike heterogeneous web videos, these samples are produced under controllable visual and motion conditions and remain close to the distribution of the foundation model being adapted. The resulting corpus therefore provides a targeted source of training data for long-horizon audio-driven avatar generation.

\noindent\textbf{Quality-aware Filtering.}
Although synthetic generation improves controllability, individual samples may still exhibit semantic mismatch, visual artifacts, static content, or camera-dominated motion. We therefore filter the generated videos along three complementary dimensions. We assess semantic consistency using multimodal similarity and reward models, including ImageBind~\cite{girdhar2023imagebind}, CLAP~\cite{laionclap2023}, and Unified Reward Model~\cite{unifiedreward}, and use Gemini~\cite{googledeepmind2026gemini35flash} to remove visually degraded videos. To assess motion quality, we compare adjacent frames to identify clips with negligible temporal change or motion dominated by simple global transformations. In particular, we reject nearly static videos and videos whose frame-to-frame variation is primarily produced by global affine camera motion, such as uniform vertical camera movement, translation, or zooming, rather than local facial and body dynamics. For dialogue-driven samples, we retain only those videos with sufficient local non-rigid motion, including visible facial articulation or body movement. This procedure yields a curated synthetic corpus with semantic consistency, visual fidelity, and motion diversity for long-horizon adaptation.

\section{Experiments}
\label{sec:experiments}

\subsection{Experimental Setup}
\label{sec:exp_setup}

\noindent\textbf{Training Data.}
We train the long-horizon adaptation branch using the fully synthetic pipeline described in Section~\ref{sec:data_pipeline}. Each sample contains an avatar video, paired audio, and a text prompt. Videos are partitioned into context-target pairs: the model conditions on a context window and predicts a video-only target chunk under the full audio condition. Unless otherwise stated, we use four latent frames as context and supervise the subsequent four latent frames. To improve robustness to temporal misalignment, the context start is sampled either from the first latent frame or from a randomly selected later position.

\noindent\textbf{Implementation Details.}
Avatar-Forever is built on the 22B LTX-2.3~\cite{hacohen2026ltx}. The efficiency branch distills the base model into a four-step generator using DMD~\cite{yin2024dmd}, while the robustness branch independently trains video-side LoRA~\cite{hu2022lora} adapters with rank and alpha both set to 128. For RRT, we perturb the historical context, roll out $K=4$ autoregressive chunks, and apply the flow-matching loss to the following target chunk. Rollout uses the default 30-step denoising schedule without CFG. History degradation is applied with probability $0.5$, including additive noise, blur, saturation, and latent masking. We always condition on the first frame through a zero-initialized gated module applied only to target denoising tokens. Both branches are optimized with AdamW at a learning rate of $1\times10^{-5}$ and a global batch size of 256; DMD distillation and RRT training run for 5,000 and 3,000 steps, respectively.

\noindent\textbf{Evaluation Protocol.}
We evaluate short-horizon quality and long-horizon stability on TalkVid~\cite{chen2026talkvid}, EMTD~\cite{meng2024echomimicv2}, and HDTF~\cite{zhang2021flow}. We construct a 5-second split for short-horizon quality and a 30-second split for long-horizon stability, each containing 40 samples with paired audio, reference identity, and text conditions. Unless otherwise specified, ablations are conducted on EMTD. We compare with representative audio-driven and long-horizon avatar methods, including OmniAvatar~\cite{gan2025omniavatar}, LiveAvatar~\cite{huang2025liveavatar}, SoulX-FlashTalk~\cite{shen2026soulxflashtalk}, and InfiniteTalk~\cite{yang2025infinitetalk}. For each baseline, we follow the official inference configuration whenever available and use identical audio and reference inputs for comparison.

\noindent\textbf{Evaluation Metrics.}
We evaluate visual quality, motion realism, audio--visual synchronization, and video-distribution similarity. For perceptual evaluation, we use \texttt{Gemini-Flash-3.5}~\cite{googledeepmind2026gemini35flash} as a strict multimodal judge, which scores audio--visual consistency, visual quality, and motion naturalness on a $1$--$5$ scale. The final score is the weighted average of them with weights $0.35$, $0.35$, and $0.30$, respectively. The complete prompt is provided in \textbf{Appendix \ref{sec:app_1}}. We additionally report Fr\'echet Inception Distance (FID), Fr\'echet Video Distance (FVD), Q-Align image quality (IQA), Q-Align aesthetic score (ASE), and Sync-C/Sync-D for lip synchronization, where higher Sync-C and lower Sync-D are better. We further conduct a user study on the EMTD long-horizon split with 20 participants. Participants rate anonymized and randomly ordered videos from Avatar-Forever and competing methods using the same three perceptual criteria as the LLM judge. We report the mean participant score, with randomization and anonymization used to reduce ordering and model-identity bias.

\subsection{Main Results}
\label{sec:main_results}

\begin{table*}[t]
\caption{
Quantitative comparison on the short- and long-video evaluation splits.
Each entry reports the 5-second / 30-second results.
Latency is measured in seconds. `w/ FC' means Avatar-Forever with ForeverCache.  
The best and second-best results are marked in bold and underline, respectively.
}
\vspace{-2mm}
\label{tab:quantitative-5s-30s}
\centering
\setlength{\tabcolsep}{3pt}
\resizebox{\textwidth}{!}{
\begin{tabular}{c|c|cccc|cccccc|c}
\toprule
\multirow{2}{*}{Dataset} & \multirow{2}{*}{Model}
& \multicolumn{4}{c|}{LLM Judge}
& \multicolumn{6}{c|}{Automatic Metrics}
& \multirow{2}{*}{Latency (s)$\downarrow$} \\
& & A-V$\uparrow$ & Visual$\uparrow$ & Motion$\uparrow$ & Overall$\uparrow$
& IQA$\uparrow$ & ASE$\uparrow$ & Sync-C$\uparrow$ & Sync-D$\downarrow$
& FID$\downarrow$ & FVD$\downarrow$ & \\
\midrule
\multirow{6}{*}{EMTD}
& OmniAvatar
& 3.77/2.92 & 3.81/1.83 & 3.06/2.00 & 3.57/2.26
& 4.15/2.98 & 2.84/2.42 & 5.57/6.83 & 9.47/8.59
& 61.62/115.84 & 979.07/1834.79 & 850.00/$>1$h \\
& InfiniteTalk
& 3.58/4.17 & 3.71/4.00 & 2.81/3.67 & 3.39/3.96
& 4.40/4.79 & 3.11/3.64 & 6.73/6.81 & 8.00/\underline{7.95}
& 39.25/37.63 & 786.25/1125.80 & 51.50/309.20 \\
& LiveAvatar
& 3.74/4.25 & \underline{4.06}/4.08 & 3.06/\underline{4.00} & 3.65/4.12
& 4.38/4.59 & 3.07/3.57 & 6.67/6.62 & 8.45/8.08
& 39.23/61.17 & 795.87/1373.52 & 53.01/320.95 \\
& SoulX
& \underline{3.87}/4.08 & 3.97/4.08 & \underline{3.13}/3.75 & \underline{3.68}/3.98
& 4.44/4.55 & 3.14/3.45 & 6.83/\underline{6.83} & \underline{7.97}/7.96
& \underline{38.91}/\underline{33.46} & 785.20/\underline{868.38} & 25.43/125.26 \\
& \textbf{Ours w/ FC}
& 3.81/\underline{4.33} & 4.00/\underline{4.25}
& \underline{3.13}/\underline{4.00} & 3.67/\underline{4.23}
& \underline{4.77}/\underline{4.84} & \underline{3.28}/\underline{3.70}
& \underline{7.11}/6.68 & 8.39/8.01
& 48.76/34.52 & \textbf{759.78}/905.97
& \textbf{4.24}/\textbf{26.71} \\
& \textbf{Ours}
& \textbf{3.90}/\textbf{4.42} & \textbf{4.16}/\textbf{4.50}
& \textbf{3.32}/\textbf{4.08} & \textbf{3.82}/\textbf{4.32}
& \textbf{4.80}/\textbf{4.88} & \textbf{3.33}/\textbf{3.73}
& \textbf{7.56}/\textbf{6.85} & \textbf{7.95}/\textbf{7.94}
& \textbf{38.37}/\textbf{33.33} & \underline{775.11}/\textbf{858.06}
& \underline{5.24}/\underline{38.85} \\
\midrule
\multirow{6}{*}{HDTF}
& OmniAvatar
& 3.73/3.80 & 3.75/3.70 & 2.88/3.00 & 3.48/3.53
& 4.01/4.03 & 2.70/3.06 & 6.14/7.30 & 9.14/7.61
& 21.31/24.01 & \underline{390.61}/956.60 & 850.00/$>1$h \\
& InfiniteTalk
& 3.34/\underline{4.00} & 3.58/3.80 & 2.47/3.00 & 3.16/3.63
& 4.03/4.15 & 2.75/3.16 & 8.14/7.43 & \underline{7.06}/7.60
& 26.71/19.48 & 390.69/793.74 & 51.50/309.20 \\
& LiveAvatar
& \underline{3.90}/3.80 & \underline{3.95}/3.80 & 3.08/3.00 & \underline{3.67}/3.56
& 4.19/4.07 & 2.74/3.23 & 7.39/6.23 & 8.39/8.45
& 23.42/95.35 & 426.26/1675.81 & 53.01/320.95 \\
& SoulX
& 3.82/\underline{4.00} & 3.80/3.70 & 2.95/\underline{3.20} & 3.55/3.66
& 4.05/4.17 & 2.75/3.20 & 8.48/7.50 & 7.10/\underline{7.54}
& 21.14/19.41 & 401.39/1066.36 & 25.43/125.26 \\
& \textbf{Ours w/ FC}
& 3.73/\textbf{4.10} & 3.88/\underline{3.90}
& \underline{3.15}/\underline{3.20} & 3.61/\underline{3.80}
& \textbf{4.42}/\textbf{4.42} & \textbf{3.38}/\textbf{3.38}
& \underline{8.56}/\underline{7.56} & 7.06/7.56
& \textbf{20.19}/\underline{17.37} & 393.92/\underline{686.96}
& \textbf{4.24}/\textbf{26.71} \\
& \textbf{Ours}
& \textbf{3.90}/\underline{4.00} & \textbf{4.03}/\textbf{4.10}
& \textbf{3.25}/\textbf{3.40} & \textbf{3.75}/\textbf{3.82}
& \underline{4.30}/\underline{4.38} & \underline{2.89}/\underline{3.33}
& \textbf{8.68}/\textbf{7.56} & \textbf{6.94}/\textbf{7.50}
& \underline{20.91}/\textbf{17.19} & \textbf{378.17}/\textbf{593.92}
& \underline{5.24}/\underline{38.85} \\
\midrule
\multirow{6}{*}{TalkVid}
& OmniAvatar
& 3.66/4.00 & 3.71/3.80 & 2.84/3.50 & 3.43/3.78
& 3.68/4.12 & 2.41/3.13 & 4.45/6.49 & 9.77/8.94
& 48.15/64.70 & 618.67/1064.48 & 850.00/$>1$h \\
& InfiniteTalk
& 3.25/4.10 & 3.13/3.90 & 2.60/3.40 & 3.01/3.82
& 3.51/4.30 & 2.31/3.26 & 5.48/6.35 & 8.87/8.90
& 53.10/52.09 & 669.80/1074.68 & 51.50/309.20 \\
& LiveAvatar
& 3.65/4.00 & 3.58/3.90 & 2.95/3.40 & 3.41/3.79
& 3.79/4.35 & 2.48/3.35 & 4.91/6.07 & 9.11/8.64
& \underline{47.47}/70.21 & 668.96/1308.11 & 53.01/320.95 \\
& SoulX
& 3.63/4.20 & 3.65/4.20 & 2.90/3.50 & 3.42/3.99
& 3.70/4.17 & 2.45/3.20 & 5.61/\underline{6.60} & 8.89/\underline{8.24}
& 47.58/\underline{51.41} & 720.48/1066.36 & 25.43/125.26 \\
& \textbf{Ours w/ FC}
& \underline{3.78}/\textbf{4.30} & \underline{3.83}/\textbf{4.40}
& \underline{3.23}/\underline{3.80} & \underline{3.63}/\underline{4.19}
& \underline{3.90}/\underline{4.47} & \underline{2.59}/\underline{3.35}
& \underline{5.68}/6.51 & \underline{8.81}/8.29
& 49.56/53.06 & \textbf{543.87}/\underline{998.75}
& \textbf{4.24}/\textbf{26.71} \\
& \textbf{Ours}
& \textbf{3.85}/\textbf{4.30} & \textbf{3.85}/\textbf{4.40}
& \textbf{3.35}/\textbf{3.90} & \textbf{3.70}/\textbf{4.22}
& \textbf{3.96}/\textbf{4.51} & \textbf{2.62}/\textbf{3.38}
& \textbf{6.01}/\textbf{6.62} & \textbf{8.61}/\textbf{8.23}
& \textbf{47.43}/\textbf{49.82} & \underline{554.00}/\textbf{994.42}
& \underline{5.24}/\underline{38.85} \\
\bottomrule
\end{tabular}
}
\vspace{-3mm}
\end{table*}

\begin{figure}[t!]
\centering
\includegraphics[width=\textwidth]{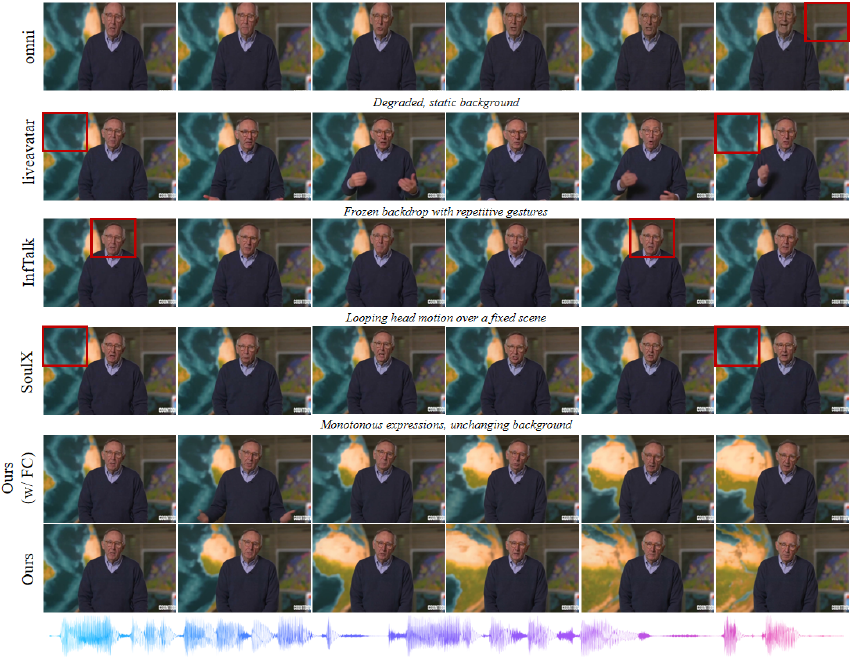}
\caption{\textbf{Visual comparison on 5-second generation.} Avatar-Forever produces more natural speech-driven facial motion, hand gestures, and background dynamics than prior methods, which often restrict motion to the face and mouth. The ForeverCache variant preserves the visual performance of the standard model while reducing inference cost.}
\centering
\label{fig:visual_short}
\end{figure}

\noindent\textbf{Short-horizon Quality.}
The 5-second results in Table~\ref{tab:quantitative-5s-30s} show that Avatar-Forever preserves the generation quality of the video foundation model under few-step inference. Compared to the strongest competing baseline, Avatar-Forever improves the LLM Overall score by $4.6\%$ on average, together with average gains of $5.1\%$ in IQA, $5.6\%$ in ASE, and $6.7\%$ in Sync-C, while consistently reducing Sync-D. The Motion score improves by up to $13.6\%$, extending the advantage beyond frame-level fidelity and synchronization to more natural speech-driven dynamics. Figure~\ref{fig:visual_short} provides qualitative comparison for these methods. Even in the short-horizon regime, the baselines already exhibit visual degradation, stereotyped facial gestures, limited motion diversity, and over-articulated mouth motion. In contrast, our results (the last two rows) preserve identity and visual fidelity while producing natural, diverse head movements and expressions with accurate lip synchronization. Beyond foreground animation, Avatar-Forever also retains coherent scene dynamics: for example, the large globe continues to rotate naturally while preserving its geographic structure and map textures, rather than becoming static or visually distorted. Please view the videos in the \textbf{\href{https://leeruibin.github.io/avatarforever-project-page/}{project page}} for better comparison. 

\begin{figure}[t!]
\centering
\includegraphics[width=\textwidth]{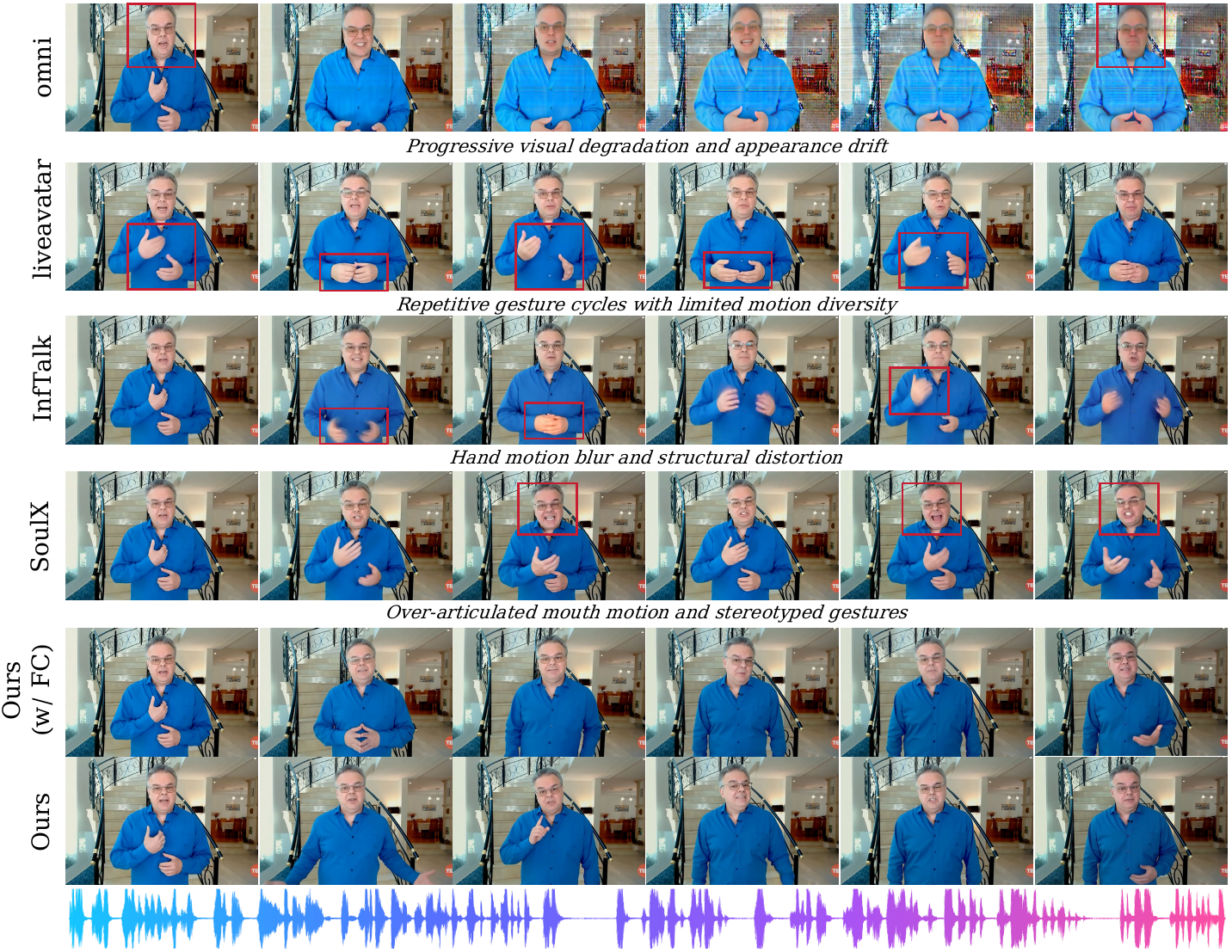}
\caption{\textbf{Visual comparison on 30-second generation.} Avatar-Forever maintains identity, facial detail, hand structure, and motion diversity over extended autoregressive rollout. In contrast, competing methods exhibit degraded hand details, repetitive gestures, or instability over time. ForeverCache retains comparable long-horizon visual stability with substantially faster inference.}
\centering
\label{fig:visual_long}
\end{figure}

\noindent\textbf{Long-horizon Stability.}
The 30-second results in Table~\ref{tab:quantitative-5s-30s} further demonstrate Avatar-Forever's robustness under autoregressive rollout. It achieves the best LLM Overall score on all three datasets, outperforming the strongest prior baseline by $5.0\%$ on average. It also improves LLM Visual quality by $7.6\%$ on average and Motion quality by up to $11.4\%$, while obtaining the best FID and FVD across all datasets. In particular, FID is reduced by $5.0\%$ on average and FVD by $25.2\%$ on HDTF, supporting improved distributional realism under accumulated context errors. Figure~\ref{fig:visual_long} provides complementary qualitative evidence for these gains. As highlighted by the red boxes, competing methods exhibit progressive visual degradation and appearance drift, repetitive low-diversity gesture cycles, motion-induced hand blur and structural distortion, or over-articulated mouth motion accompanied by stereotyped upper-body gestures. In contrast, our Avatar-Forever preserves identity, facial and hand details, and diverse natural motion throughout the extended rollout. Its ForeverCache variant retains comparable long-horizon stability despite the substantial inference acceleration. 

\noindent\textbf{Efficiency-Quality Balance.}
ForeverCache improves streaming efficiency by reusing fixed historical context features across denoising steps. On short videos, it reduces latency by $19.1\%$ and increases throughput by $23.6\%$ compared to standard Avatar-Forever, while remaining approximately $6.0\times$ faster than the fastest prior baseline. On 30-second generation, it reduces runtime from $38.85$s to $26.71$s, corresponding to a $31.2\%$ latency reduction and a $45.5\%$ throughput increase; it remains approximately $4.7\times$ faster than the fastest prior method. Despite this acceleration, ForeverCache preserves most quality gains of the standard model: it remains competitive across perceptual, synchronization, and distributional metrics on short videos, and achieves the second-best LLM Overall score on all long-video splits while outperforming the strongest prior baseline by $3.8\%$ on average. Figures~\ref{fig:visual_long} show that cache-enabled inference retains the key visual behavior of the full model without visible collapse or systematic temporal degradation.

\begin{figure*}[t!]
\centering
\includegraphics[width=\textwidth]{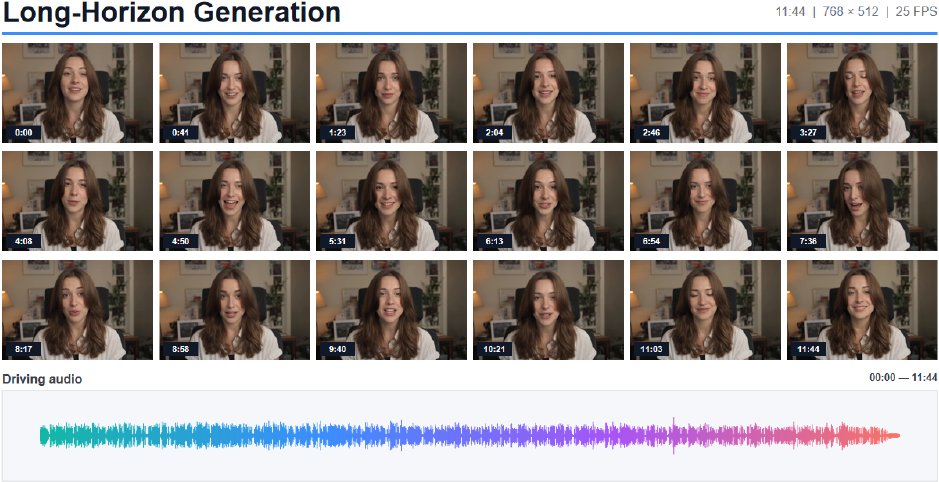}
\caption{\textbf{Visual results of extended-duration generation.} We uniformly sample frames from a continuous video generated for more than 11 minutes. Avatar-Forever maintains stable identity, facial structure, and scene content while producing natural expressions, coherent head and body motion, and accurate audio--visual synchronization throughout the extended autoregressive rollout. The bottom row shows the corresponding driving-audio waveform.}
\centering
\label{fig:vis_10min}
\end{figure*}

\noindent\textbf{Extended-Duration Generation}. Figure~\ref{fig:vis_10min} presents an audio-driven avatar video generated continuously for more than 11 minutes. Avatar-Forever maintains stable identity, facial structure, appearance, and scene content without progressive drift or visible quality collapse. The avatar exhibits natural speech-driven dynamics, including realistic facial expressions, coherent gaze, subtle head movements, smooth posture changes, and non-repetitive gestures. These motions remain temporally coherent and appropriately follow the rhythm and emphasis of the speech, avoiding the frozen expressions, mechanical motion, and repetitive behavior commonly observed in long-form generation. Moreover, lip movements and facial articulation remain consistently aligned with the driving audio throughout the rollout. This result demonstrates that Avatar-Forever preserves visual fidelity, motion naturalness, and audio--visual consistency well beyond the 30-second evaluation setting, supporting its capability for stable generation without a predefined duration. 

\vspace{+1mm}
\noindent Videos of the above examples and more visual examples can be found in \textbf{Appendix \ref{sec:app_2}} and \textbf{\href{https://leeruibin.github.io/avatarforever-project-page/}{project page}}. 
\vspace{+1mm}

\begin{table*}[t]
    \caption{
    Double-blind user study results on the EMTD long-video evaluation split.
    Each score is computed from user ratings on a 1--5 scale and normalized to a 0--100 scale.
    The best and second-best results are highlighted in bold and underlined, respectively.
    }
    \label{tab:user-study}
    \centering
    \resizebox{0.6\textwidth}{!}{
    \begin{tabular}{c|cccc}
        \toprule
        \multirow{2}{*}{Model}
        & \multicolumn{4}{c}{User Study} \\
        & Audio-Visual$\uparrow$ & Visual$\uparrow$ & Motion$\uparrow$ & Overall$\uparrow$ \\
        \midrule
        OmniAvatar      & 57.5 & 32.5 & 29.0 & 40.20 \\
        InfiniteTalk    & 68.0 & 51.5 & 45.0 & 55.33 \\
        LiveAvatar      & 70.0 & 54.5 & 47.0 & 57.68 \\
        SoulX-FlashTalk & 71.0 & 56.5 & 52.0 & 60.23 \\
        Ours w/ ForeverCache & \underline{73.0} & \underline{76.5} & \underline{75.0} & \underline{74.83} \\
        Ours            & \textbf{74.0} & \textbf{78.0} & \textbf{76.0} & \textbf{76.00} \\
        \bottomrule
    \end{tabular}
    }
\end{table*}

\noindent\textbf{Human Perceptual Validation.}
Table~\ref{tab:user-study} reports the double-blind user study results on the EMTD long-video split. Twenty participants rate anonymized videos on the same four perceptual dimensions as the LLM judge using integer scores from 1 to 5; scores are then averaged and normalized to a 0--100 scale. The human evaluation is consistent with the automatic results: Avatar-Forever ranks first on all dimensions, while the ForeverCache variant consistently ranks second with only a small gap from the full model.

The advantage in human validation is particularly pronounced for Visual and Motion quality, highlighting aspects of temporal naturalness that are not fully captured by automatic evaluation. While several baselines preserve identity and produce locally plausible frames, their long videos often exhibit limited upper-body motion, static backgrounds, repetitive gestures, rigid posture changes, or exaggerated mouth shapes. In contrast, Avatar-Forever produces more expressive facial dynamics, smoother gesture transitions, and motion that better follows the rhythm of speech. The resulting improvement in temporal naturalness also strengthens perceived visual realism, explaining the highest Overall score assigned by human raters. ForeverCache preserves most of these perceptual gains while providing substantially faster inference.

\subsection{Analysis of Decoupled Training}

\begin{table*}[t]
    \caption{
    Ablation study on decoupled training. We compare the proposed decoupled training strategy with DMD-only training and FM-only long-horizon adaptation. Best results are highlighted in bold, and second-best results are underlined.
    }
    \label{tab:ablation-decoupled-training}
    \centering
    \resizebox{\textwidth}{!}{
    \begin{tabular}{c|cccc|cccccc}
        \toprule
        \multirow{2}{*}{Setting}
        & \multicolumn{4}{c|}{LLM Judge}
        & \multicolumn{6}{c}{Automatic Metrics} \\
        & Audio-Visual$\uparrow$ & Visual$\uparrow$ & Motion$\uparrow$ & Overall$\uparrow$
        & IQA$\uparrow$ & ASE$\uparrow$ & Sync-C$\uparrow$ & Sync-D$\downarrow$ & FID$\downarrow$ & FVD$\downarrow$ \\
        \midrule
        Decoupled DMD + RRT
        & \textbf{4.250} & \textbf{4.267} & \textbf{3.750} & \textbf{4.105}
        & \textbf{4.851} & \textbf{3.684} & \textbf{6.694} & \textbf{8.032} & \textbf{34.978} & \textbf{901.811} \\
        
        DMD only
        & \underline{4.083} & \underline{4.167} & \underline{3.583} & \underline{3.962}
        & \underline{4.793} & \underline{3.671} & \underline{6.427} & \underline{8.085} & \underline{39.515} & 1080.230 \\
        
        FM only
        & 2.750 & 1.917 & 2.250 & 2.308
        & 3.409 & 2.590 & 2.031 & 10.760 & 76.577 & \underline{1071.200} \\
        \bottomrule
    \end{tabular}
    }
\end{table*}

\begin{figure}[t!]
\centering
\includegraphics[width=\textwidth]{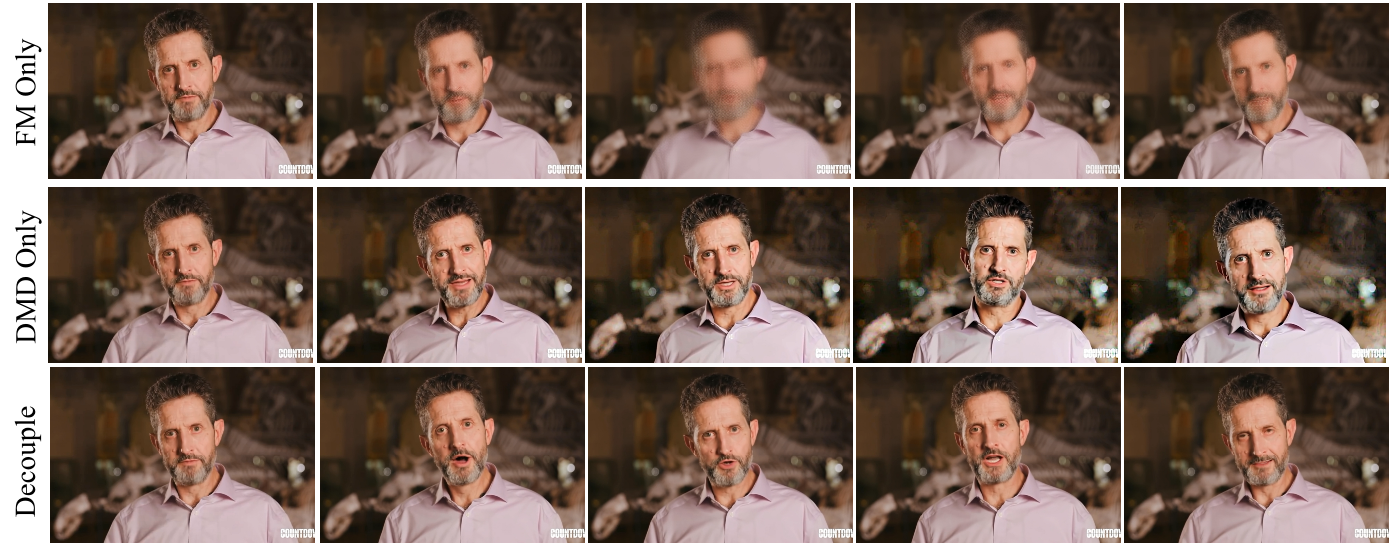}
\caption{\textbf{Effect of decoupled training.} DMD only preserves short-horizon appearance but is susceptible to long-horizon drift, whereas FM-only adaptation degrades visual fidelity. Combining DMD with RRT preserves detailed appearance while improving stability under autoregressive generation.}
\centering
\label{fig:able_branch}
\end{figure}

Table~\ref{tab:ablation-decoupled-training} and Figure~\ref{fig:able_branch} test our central hypothesis that few-step efficiency and long-horizon robustness are distinct capabilities. The full Decoupled DMD + RRT model achieves the best result on every metric, spanning perceptual quality, motion, synchronization, and distributional realism. Relative to DMD only, adding RRT improves the LLM Overall score by $3.6\%$, increases Sync-C by $4.2\%$, and reduces FID and FVD by $11.5\%$ and $16.5\%$, respectively. Although DMD provides a strong few-step generator, it is not exposed to the accumulated self-conditioning errors encountered during autoregressive inference. Correspondingly, Figure~\ref{fig:able_branch} shows weaker long-horizon appearance stability.

The FM only variant also shows limited performance. Optimizing long-horizon flow matching without the distilled generator substantially degrades visual fidelity and synchronization, producing blurred facial details and poor perceptual quality. Compared with this variant, the full model improves LLM Overall by $77.9\%$, increases Sync-C by $229.6\%$, and reduces FID by $54.3\%$. These results show that robustness cannot be obtained by optimizing long-horizon recovery alone. DMD preserves the few-step generation prior, whereas RRT adapts it to the rollout-time error distribution; independently learning and composing these capabilities yields both high local fidelity and stable long-horizon generation.

\subsection{Effect of Recovery-oriented Rollout Training}

\begin{figure}[t!]
\centering
\includegraphics[width=\textwidth]{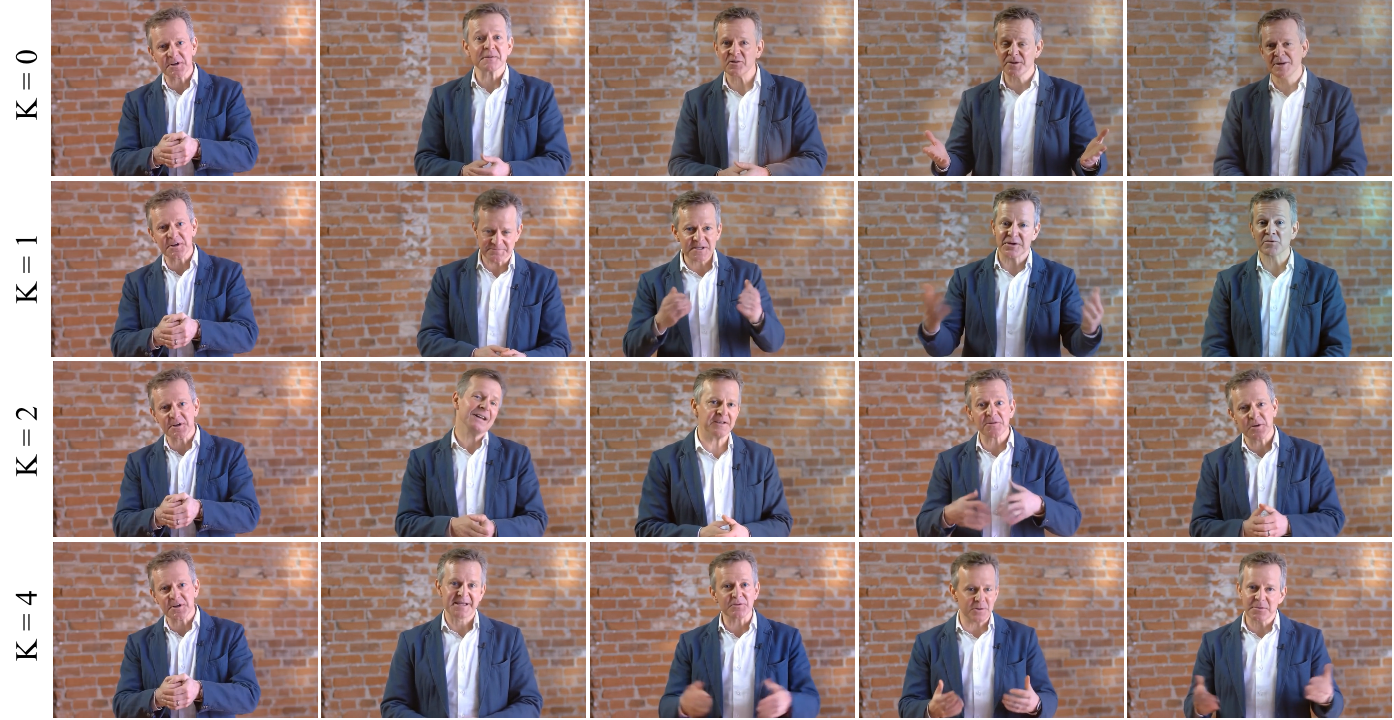}
\caption{\textbf{Effect of the RRT rollout horizon.} Immediate supervision after context degradation ($K=0$) leaves visible artifacts and appearance drift. Increasing the autoregressive rollout horizon exposes the model to accumulated self-generated errors before supervision, improving color, background, and identity consistency; $K=4$ yields the most stable result.}
\centering
\label{fig:able_RRT}
\end{figure}

Figure~\ref{fig:able_RRT} isolates the effect of the rollout horizon $K$ in RRT. The key observation is that corrupted-history training alone is insufficient: when $K=0$, supervision is applied immediately after context perturbation, reducing the task to local reconstruction. The resulting videos remain susceptible to accumulated autoregressive errors, exhibiting spot-like artifacts and global color drift. Increasing the horizon to $K=1$ partially exposes the model to self-generated errors, but visible appearance fluctuations remain.

Larger rollout horizons  match more closely the error accumulation process encountered at inference time. Both $K=2$ and $K=4$ improve appearance, background, and color consistency, with $K=4$ producing the most stable long-video results. This progression supports the design of RRT: its benefit comes from supervising recovery after model-induced errors have propagated through rollout, rather than from merely injecting synthetic corruption into historical context.

\section{Conclusion}

We presented Avatar-Forever, a framework for real-time long-duration audio-driven avatar generation. Our key insight was that the quality of few-step generation and the robustness to autoregressive drift are distinct adaptation objectives. We therefore learned them independently: DMD produced an efficient few-step generator, RRT trained a lightweight adapter to recover from accumulated rollout errors, and ForeverCache reduced redundant historical computation during streaming inference. Built on a 22B video foundation model and a fully synthetic data pipeline, Avatar-Forever improved perceptual quality, audio--visual synchronization, long-horizon stability, and inference efficiency. These results supported a simple principle for scalable video adaptation: learn efficiency and long-horizon robustness separately, then compose them at deployment. 

\noindent\textbf{Limitations and Future Work.}
Avatar-Forever achieves 27.2 FPS at $768\times512$ resolution on a single H100 GPU, but it is not yet optimized for consumer-grade hardware. In addition, although our training and optimization are tailored to audio-driven avatars, we observe a good generalization to broader video-generation scenarios, as illustrated in our \textbf{\href{https://leeruibin.github.io/avatarforever-project-page/}{project page}}. Future work will explore domain-specific data construction and training strategies to further unlock Avatar-Forever's potential for general long-horizon video generation.

\section*{Ethics Statement}

This work aims to advance long-duration audio-driven avatar generation for constructive applications such as virtual communication, education, accessibility, and digital content creation. The evaluation data used in this paper are drawn from publicly available and openly accessible research datasets. Our synthetic training data are generated from the underlying video foundation model and filtered for quality, motion, and semantic consistency, rather than collected from private or restricted personal sources. We recognize that high-fidelity avatar synthesis is a dual-use technology. While it can support beneficial interactive media applications, it may also be misused for impersonation, deceptive content generation, or misinformation. We do not endorse any use of this technology to create unauthorized representations of real individuals, manipulate public opinion, or violate privacy, consent, or intellectual property rights. 

To reduce potential misuse, we encourage deployment together with responsible safeguards, including identity-consent verification, synthetic-content disclosure, invisible watermarking, provenance tracking, and robust forgery detection. We also recommend that generated videos be clearly labeled as synthetic when used in public-facing scenarios. Our goal is to contribute to safer and more transparent digital human generation, and we will continue to align future development with responsible AI principles.


{\small
\bibliography{ref}
}

\appendix

\clearpage
\setcounter{table}{0}
\setcounter{equation}{0}
\setcounter{figure}{0}
\renewcommand{\thetable}{\thesection.\arabic{table}}
\renewcommand{\theequation}{\thesection.\arabic{equation}}
\renewcommand{\thefigure}{\thesection.\arabic{figure}}
\begin{center}
    {\LARGE\bfseries Appendix}
\end{center}

In this appendix, we provide the LLM-based perceptual evaluation prompt, and more qualitative results to complement the main paper. Specifically, the supplement includes:

\begin{enumerate}[label=\textbf{\Alph*.}, leftmargin=2em]
    \item The LLM-based perceptual evaluation prompt (referring to Sec. \ref{sec:exp_setup} in the main paper);
    \item Additional visual comparisons for 5-second and 30-second audio-driven avatar generation, including results with and without ForeverCache (referring to Sec. \ref{sec:main_results} in the main paper).
\end{enumerate}

For better viewing experience, we uploaded all the videos to our project page  \textbf{\href{https://leeruibin.github.io/avatarforever-project-page/}{project page}}, where the videos can be played directly in the browser.
Note that, due to the significant number of high-quality video files included in our demonstrations, initial page loading may require several minutes to complete. We appreciate your patience during this process, as the complete visual experience is essential to understand the capabilities and performance of our approach.

\section{LLM-based Perceptual Evaluation Prompt}
\label{sec:app_1}

Conventional automatic metrics measure image quality, video realism, and audio--visual synchronization, but they do not fully capture whether an avatar behaves like a natural speaker throughout a continuous video. In particular, temporally accumulated artifacts, repetitive gestures, frozen expressions, and unnatural motion transitions can be difficult to characterize using individual automatic metrics. We therefore employ \texttt{Gemini-Flash-3.5}~\cite{googledeepmind2026gemini35flash} as a multimodal perceptual judge. For each generated sample, the evaluator receives only the corresponding audio and video to assess the result using a fixed evaluation rubric shared across all methods, datasets, and video durations.

The evaluation covers three complementary dimensions. \emph{Audio--Visual Consistency} measures whether mouth shapes, facial dynamics, head motion, speaking rhythm, and pauses are synchronized with the input audio. \emph{Visual Quality} evaluates facial fidelity, sharpness, identity consistency, lighting, texture realism, background stability, and temporal artifacts such as flickering, blur, and appearance drift. \emph{Motion Naturalness} focuses on the temporal behavior of the avatar, including facial expressions, blinking, head and upper-body movement, breathing, gesture diversity, contextual appropriateness, and physical plausibility.

Each dimension is assigned an integer score from 1 to 5 according to explicit quality anchors, with a score of 5 reserved for videos that are nearly indistinguishable from high-quality real talking videos. The overall perceptual score is computed as:
\begin{equation}
S_{\mathrm{overall}}
=
0.35S_{\mathrm{A\text{-}V}}
+
0.35S_{\mathrm{visual}}
+
0.30S_{\mathrm{motion}},
\end{equation}
and rounded to two decimal places. The evaluator additionally returns a brief justification for each score in a structured JSON format, enabling consistent parsing and inspection of the judgments.

The prompt is designed to be particularly sensitive to long-horizon generation failures. It explicitly penalizes accumulated blur, identity or appearance drift, visual artifacts, repetitive gesture cycles, and motion that gradually becomes frozen or mechanical. It also prevents large but inappropriate motion from being rewarded solely for its magnitude. To avoid evaluation based on irrelevant personal attributes, the judge is instructed not to consider the identity, gender, ethnicity, or appearance of the depicted person. The complete system prompt used for evaluation are provided in Figure \ref{fig:llm-evaluation-prompt}.

\begin{figure*}[!t]
\centering
\begin{adjustbox}{width=0.8\textwidth,center}
\begin{minipage}{\textwidth}
\hrule
\begin{Verbatim}[
fontsize=\scriptsize,
breaklines=true,
breakanywhere=true
]
You are an expert evaluator for audio-driven talking avatar videos. Your task is to evaluate the quality of a generated avatar video based only on the provided audio and video. Please act as a strict but fair human judge.
Evaluate the video from the following three aspects:
Audio-Visual Consistency / Lip-Sync Quality
Assess whether the facial movements, mouth shapes, lip motion, head motion, and speaking rhythm are well synchronized with the input audio.5: The mouth movements and facial dynamics are highly synchronized with the audio. Lip shapes, speaking rhythm, pauses, and expressions are natural and convincing.
4: Mostly synchronized, with only minor lip-sync or timing errors.
3: Roughly synchronized, but noticeable mismatches exist in mouth motion, rhythm, or expression.
2: Poor synchronization. The mouth often moves at the wrong time or does not match the speech well.
1: Severe mismatch between audio and video. The avatar appears unrelated to the audio.

Visual Quality
Assess the overall visual fidelity of the video, including face quality, sharpness, identity consistency, lighting, texture realism, background stability, and absence of artifacts.5: High visual quality with realistic face details, stable identity, clean textures, natural lighting, and almost no artifacts.
4: Good visual quality with minor artifacts, slight blur, or small temporal inconsistencies.
3: Acceptable quality, but with visible artifacts, flickering, blur, identity drift, or unnatural textures.
2: Low quality with strong artifacts, unstable face, obvious distortion, or degraded background.
1: Very poor visual quality with severe artifacts, broken face structure, or unusable video.

Motion Naturalness / Content Naturalness
Assess whether the avatar behaves naturally over time. Focus on head movement, blinking, facial expressions, body gestures, breathing, and whether the motion is diverse and contextually appropriate.5: The avatar shows natural, smooth, and diverse motion. Expressions, blinking, head movement, and small gestures are realistic and not repetitive.
4: Mostly natural motion, with only slight stiffness or occasional repetition.
3: Somewhat natural, but the motion is limited, mechanical, repetitive, or lacks expressive variation.
2: Clearly unnatural motion. The avatar repeats the same gestures, has frozen expressions, or shows robotic movement.
1: Extremely unnatural. The avatar is almost static, heavily repetitive, jittery, or physically implausible.

Important evaluation rules:
Do not judge based on the identity, gender, ethnicity, or appearance of the person.
Do not reward a video simply because it has large motion; motion should be natural and appropriate.
Penalize repetitive behaviors, such as repeatedly nodding, blinking, smiling, moving hands, or making the same head motion without semantic variation.
Penalize long-term degradation, including identity drift, accumulated blur, increasing artifacts, or motion becoming frozen/repetitive over time.
If the video contains a speaking avatar, prioritize whether the avatar looks like a natural person speaking the given audio.
Be strict. A score of 5 should only be given when the video is nearly indistinguishable from a high-quality real talking video.
Please output your evaluation in the following JSON format only:
{
"audio_visual_consistency": {
"score": <integer from 1 to 5>,
"reason": "<brief explanation>"
},
"visual_quality": {
"score": <integer from 1 to 5>,
"reason": "<brief explanation>"
},
"motion_naturalness": {
"score": <integer from 1 to 5>,
"reason": "<brief explanation>"
},
"overall_score": <weighted average score from 1 to 5>,
"overall_reason": "<brief final judgment>"
}
Compute the overall_score using the following weights:
Audio-Visual Consistency: 0.35
Visual Quality: 0.35
Motion Naturalness: 0.30
Round the overall_score to two decimal places.
\end{Verbatim}
\hrule
\end{minipage}
\end{adjustbox}
\caption{\textbf{System prompt for LLM-based perceptual evaluation.}
The same evaluation prompt is applied to all methods, datasets, and
video durations. It defines the scoring criteria for audio--visual
consistency, visual quality, and motion naturalness, as well as the
weighted aggregation used to compute the overall score.}
\label{fig:llm-evaluation-prompt}
\end{figure*}

\section{Visual Results and Comparisons}
\label{sec:app_2}

\begin{figure*}[t!]
\centering
\includegraphics[width=\textwidth]{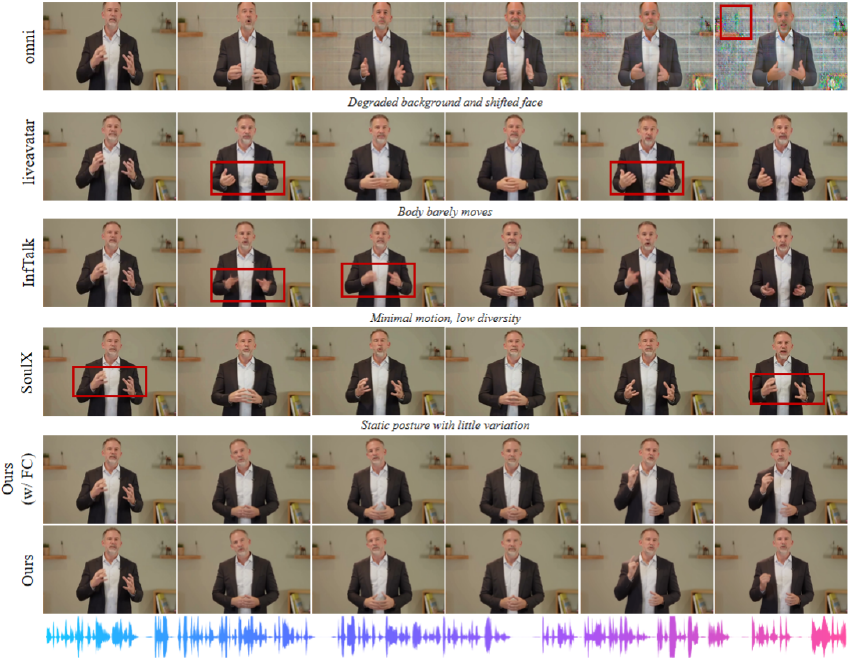}
\vspace{-2mm}
\caption{\textbf{Visual comparison on 30-second generation (sample 1).} From top to bottom, the rows show OmniAvatar, LiveAvatar, InfiniteTalk, SoulX-FlashTalk, \emph{Ours (w/ FC)}, and \emph{Ours}; the columns show frames uniformly sampled over time, with the driving audio shown below. The baselines drift with a degraded background and a shifted face, and the body barely moves. Avatar-Forever (last two rows) keeps identity and background stable while the head and body lean and rotate for more vivid, diverse motion. ForeverCache retains comparable visual stability with substantially faster inference.}
\centering
\label{fig:vis_long_case1}
\end{figure*}

\begin{figure*}[t!]
\centering
\includegraphics[width=\textwidth]{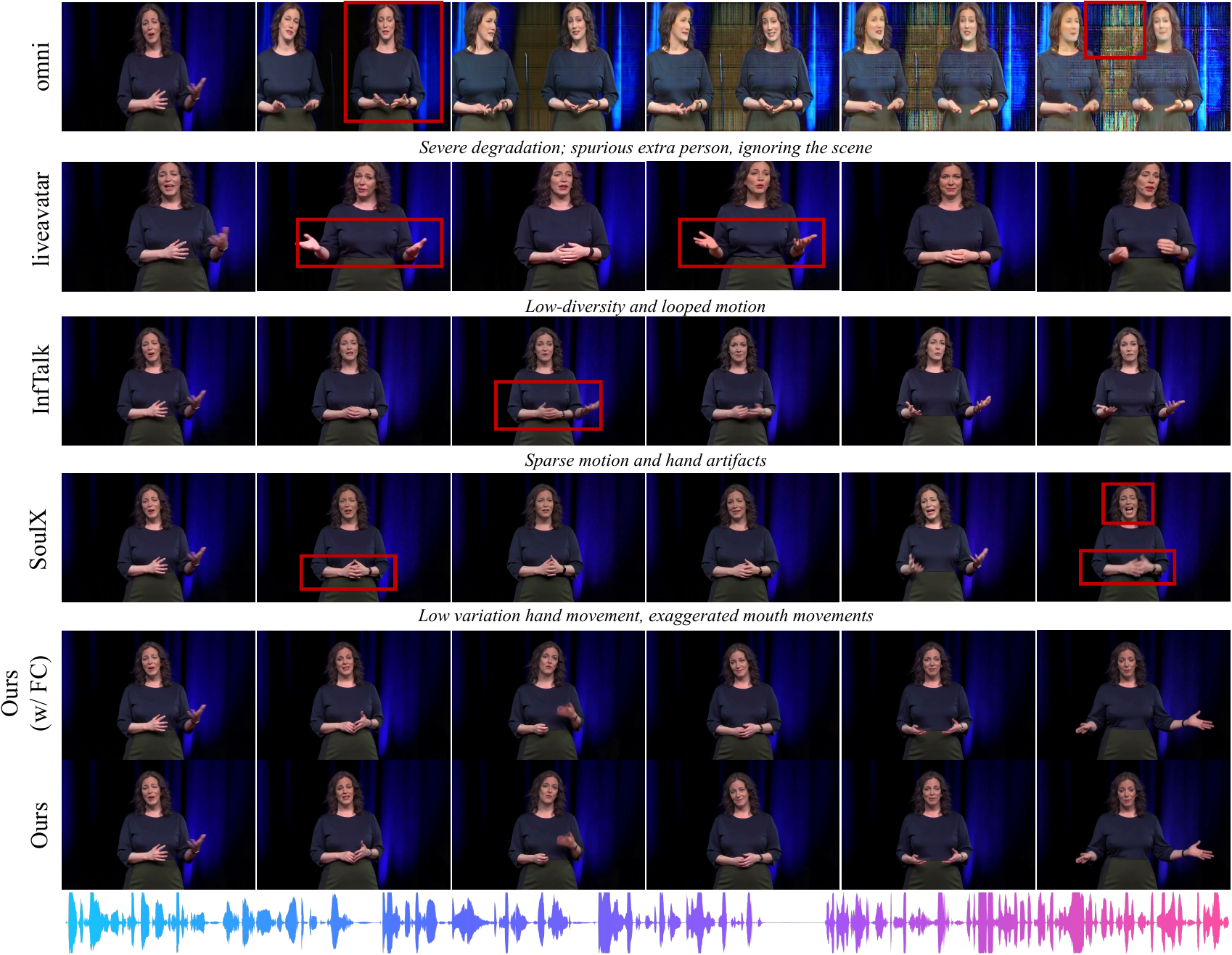}
\vspace{-2mm}
\caption{\textbf{Visual comparison on 30-second generation (sample 2).} Same row and column layout as Figure~\ref{fig:vis_long_case1}. The baselines show severe degradation, including a spurious extra person that ignores the scene, together with low motion diversity. Avatar-Forever (last two rows) remains stable over the full 30 seconds with larger, more diverse hand and arm motion. ForeverCache retains comparable visual stability with substantially faster inference.}
\centering
\label{fig:vis_long_case13}

\end{figure*}

\noindent\textbf{30s Long-Video Generation}. 
Figures~\ref{fig:vis_long_case1}--\ref{fig:vis_long_case13} compare 30-second results on two long-horizon samples. In each figure, from top to bottom, the rows show OmniAvatar, LiveAvatar, InfiniteTalk, SoulX-FlashTalk, and our method with and without ForeverCache (\emph{Ours (w/ FC)} and \emph{Ours}); the columns show uniformly sampled frames from start to end, and the bottom strip visualizes the driving audio. Over the longer horizon, the baselines accumulate errors that are far more pronounced than in the short setting: background and appearance drift, severe visual degradation and spurious content (e.g., an extra person that ignores the scene), skin-color shift, and motion that is either nearly static or dominated by camera movement. Our results (the last two rows) remain stable across the full duration, keeping the background and identity consistent while generating more vivid and diverse head and body motion, such as the head leaning and rotating rather than repeating a fixed pose. At the same time, AvatarForever with ForeverCache closely matches the base AvatarForever, confirming that ForeverCache preserves long-horizon quality while substantially reducing redundant history computation.

\begin{figure*}[t!]
\centering
\includegraphics[width=\textwidth]{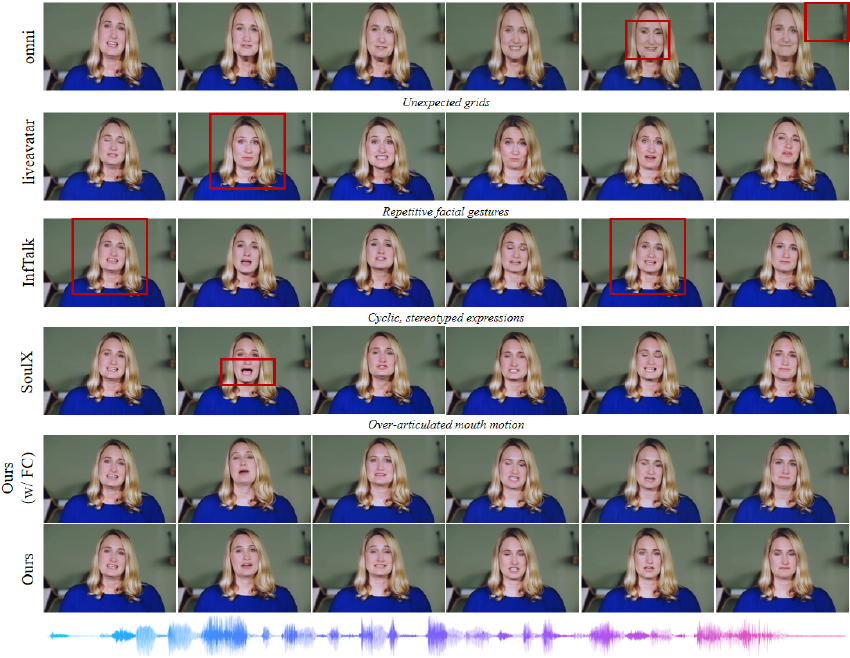}
\vspace{-2mm}
\caption{\textbf{Visual comparison on 5-second generation (sample 1).} From top to bottom, the rows show OmniAvatar, LiveAvatar, InfiniteTalk, SoulX-FlashTalk, \emph{Ours (w/ FC)}, and \emph{Ours}; the columns show frames uniformly sampled over time, with the driving audio shown below. The baselines show unexpected grids, repetitive and stereotyped facial gestures, and over-articulated mouth motion. Avatar-Forever (last two rows) preserves identity and facial detail with more natural, diverse expressions and accurate lip synchronization. ForeverCache retains comparable visual quality with substantially faster inference.}
\centering
\label{fig:vis_short_1}
\end{figure*}

\begin{figure*}[t!]
\centering
\includegraphics[width=\textwidth]{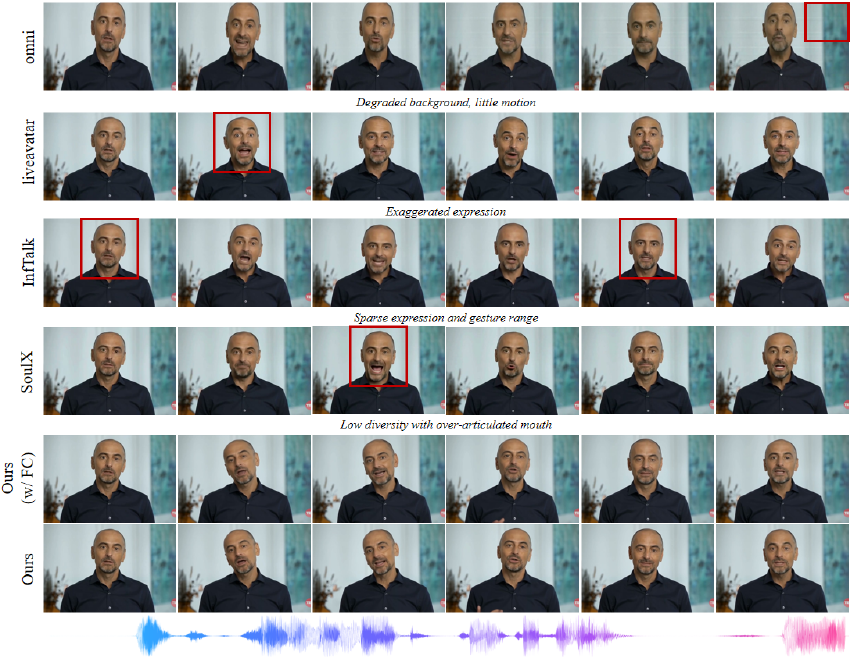}
\vspace{-2mm}
\caption{\textbf{Visual comparison on 5-second generation (sample 2).} Same row and column layout as Figure~\ref{fig:vis_short_1}. The baselines exhibit background degradation, over-articulated expression and mouth motion. Avatar-Forever (last two rows) achieves richer motion diversity, such as the head leaning and turning, while keeping a stable appearance. ForeverCache retains comparable visual quality with substantially faster inference.}
\centering
\label{fig:vis_short_2}
\end{figure*}

\noindent\textbf{5s Short-Video Generation}. 
Figures~\ref{fig:vis_short_1}--\ref{fig:vis_short_2} compare 5-second results on two EMTD samples. Even in the short-horizon regime, the baselines already exhibit visual degradation, repetitive and stereotyped facial gestures, limited motion diversity, and over-articulated mouth motion. In contrast, our results (the last two rows) preserve identity and visual fidelity while producing more natural and diverse head motion and expressions with accurate lip synchronization. AvatarForever with ForeverCache is visually on par with base AvatarForever, showing that ForeverCache accelerates streaming inference without sacrificing quality.

\end{document}